\documentclass{article}
\usepackage[accepted]{icml2022}
\usepackage{microtype}
\usepackage{amsmath,graphicx}
\usepackage{amssymb,amsfonts,amsthm}
\usepackage{mathrsfs}
\usepackage{mathtools}
\usepackage[mathscr]{eucal}
\usepackage{bbm}
\usepackage{bm}
\usepackage{times}

\usepackage{xcolor}
\usepackage{url}
\usepackage{hyperref}
\usepackage[capitalize,noabbrev]{cleveref}
\hypersetup{
    colorlinks,
    linkcolor={red!75!black},
    citecolor={blue!80!black},
    urlcolor={blue!80!black}
}
\usepackage{booktabs}
\usepackage{subcaption}

\usepackage{multicol}
\usepackage{multirow}
\usepackage{paralist, enumitem}

\usepackage{algorithm}
\usepackage{algorithmic}
\usepackage[titlenumbered,linesnumbered,ruled,noend,algo2e]{algorithm2e}

\SetCommentSty{mycommfont}
\SetEndCharOfAlgoLine{}

\usepackage{nicefrac}
\usepackage{float}
\usepackage{wrapfig}

\usepackage{thmtools, thm-restate}
\theoremstyle{plain}
\newtheorem{theorem}{Theorem}[section]
\newtheorem{proposition}[theorem]{Proposition}
\newtheorem*{proposition*}{Proposition}
\newtheorem{lemma}[theorem]{Lemma}

\theoremstyle{definition}
\newtheorem{definition}[theorem]{Definition}

\theoremstyle{remark}

\Crefname{proposition}{Proposition}{Propositions}
\crefname{problem}{Problem}{Problems}
\Crefname{lemma}{Lemma}{Lemmas}

\usepackage{aliascnt}
\newaliascnt{problem}{equation}
\aliascntresetthe{problem}
\creflabelformat{problem}{#2\textup{#1}#3}

\makeatletter

\def\endproblem{\eqno \hbox{\@eqnnum}$$\@ignoretrue}
\makeatother

\crefname{problem}{Problem}{Problems}
\crefname{algorithm}{Algorithm}{Algorithms}
\crefname{figure}{Figure}{Figures}
\crefname{proposition}{Proposition}{Propositions}
\crefname{appendix}{Appendix}{Appendix}

\newcommand{\al}[1]{\textcolor{teal}{#1}}
\newcommand{\z}[1]{\textcolor{purple}{\textbf{Zoltan:} #1}}

\newcommand{\iid}{{i.i.d.~}}

\newcommand{\eg}{{\em e.g.~}}

\newcommand{\st}{{s.t.~}}

\newcommand{\rvX}{\mathsf{X}}
\newcommand{\rvY}{\mathsf{Y}}

\newcommand{\mH}{\mathscr{H}}

\newcommand{\mL}{\mathscr{L}}

\newcommand{\mS}{\mathscr{S}}

\newcommand{\mX}{\mathscr{X}}
\newcommand{\mY}{\mathscr{Y}}

\newcommand{\cB}{\mathcal{B}}
\newcommand{\cC}{\mathcal{C}}

\renewcommand{\b}{\mathbf}
\newcommand{\tb}{\textbf}

\newcommand{\bbN}{\mathbb{N}}

\newcommand\R{\mathbb{R}}
\newcommand\reals{\mathbb{R}}
\DeclareMathOperator{\dom}{dom}

\DeclareMathOperator{\prox}{prox}
\DeclareMathOperator{\Proj}{Proj}

\DeclareMathOperator{\sign}{sign}
\DeclareMathOperator{\Id}{Id}
\DeclareMathOperator{\Span}{Span}

\DeclareMathOperator{\Tr}{Tr}
\DeclareMathOperator{\infconv}{\square}

\newcommand{\scalone}[1]{\langle \rangle}

\newcommand{\norm}[1]{\left\lVert {#1} \right\rVert}

\newcommand{\abs}[1]{\left\lvert #1 \right\rvert}

\newcommand{\argmin}{\mathop{\mathrm{arg\,min}}}

\newcommand{\condset}[2]{ \left\{ #1 \,:\, #2 \right\} } 

\renewcommand{\d}{\mathrm{d}} 

\icmltitlerunning{Functional Output Regression with Infimal Convolution}

\begin{document}

\twocolumn[
\icmltitle{Functional Output Regression with Infimal Convolution: \\
Exploring the Huber and $\epsilon$-insensitive Losses}



\icmlsetsymbol{equal}{*}

\begin{icmlauthorlist}
\icmlauthor{Alex Lambert}{tpt,kul}
\icmlauthor{Dimitri Bouche}{tpt}
\icmlauthor{Zolt{\'a}n Szab{\'o}}{lse}
\icmlauthor{Florence d'Alch\'{e}-Buc}{tpt}
\end{icmlauthorlist}

\icmlaffiliation{kul}{ESAT, KU Leuven, Belgium}
\icmlaffiliation{tpt}{LTCI, Telecom Paris, IP Paris, France}
\icmlaffiliation{lse}{Department of Statistics, London School of Economics, United Kingdom}

\icmlcorrespondingauthor{Alex Lambert}{alex.lambert@kuleuven.be}

\icmlkeywords{kernel methods, functional data}

\vskip 0.3in
]



\printAffiliationsAndNotice{}  

\begin{abstract}
  The focus of the paper is functional output regression (FOR) with convoluted losses.
  While most existing work consider the square loss setting,
  we leverage extensions of the Huber and the $\epsilon$-insensitive loss (induced by infimal convolution) and propose a flexible framework capable of handling various forms of outliers and sparsity in the FOR family.
  We derive computationally tractable algorithms relying on duality to tackle the resulting tasks in the context of vector-valued reproducing kernel Hilbert spaces.
  The efficiency of the approach is demonstrated and contrasted with the classical squared loss setting on both synthetic and real-world benchmarks.
\end{abstract}
\section{Introduction}
\label{sec:introduction}
Functional data analysis (FDA, \citealt{Ramsay97functionaldata, wang2016functional}) has attracted a growing attention in the field of machine learning and statistics, with applications for instance in biomedical signal processing \citep{ullah2013applications}, epidemiology monitoring and climate science \citep{ramsay2007applied}.
The key assumption is that we have access to densely-measured observations, in which case functional data description becomes the most natural and adequate.
An important subfield of FDA is functional output regression (FOR) which focuses on regression problems where the output variable is a function. 
There are numerous ways to tackle the FOR problem family.
The simplest approach is to assume linear dependence between the inputs and the outputs \citep{morris2015functional}.
However, in order to cope with more complex dependencies, various nonlinear approaches have been designed.
In nonparametric statistics, \citet{ferraty2011kernel} proposed a Banach-valued Nadaraya-Watson estimator.
The flexibility of kernel methods \citep{steinwart2008support} and the richness of the associated reproducing kernel Hilbert spaces (RKHSs; \citealt{micchelli06universal}) have proven to be particularly useful in the area, with works involving tri-variate regression problem \citep{reimherr2018optimal}, and approximated kernel ridge regression (KRR) using orthonormal bases \citep{oliva15fast}.
In the operator-valued kernel \citep{Pedrick57, carmeli2010vector} literature, examples include function-valued KRR with double representer theorem \citep{lian2007nonlinear}, solvers based on the discretization of the loss function \citep{Kadri2010}, purely functional methods relying on approximate inversion of integral operators \citep{kadri2016ovk} or techniques relying on finite-dimensional coefficients of the functional outputs in a dictionary basis \citep{bouche21a}.

Most of these works employ the square loss which induces an estimate of the conditional expectation of the functional outputs given the input data.
However defective sensors or malicious attacks can lead to erroneous or contaminated measurements \citep{hubert15outliers}, resulting in local or global functional outliers.
The square loss is expected to be badly affected in those cases and considering alternative losses is a natural way to obtain reliable and robust prediction systems.
For scalar-valued outputs, the Huber loss \citep{huber1964robust} and the $\epsilon$-insensitive loss \citep{lee2005epsilon} are particularly
popular and well-suited to construct outlier-robust estimators.
In the FDA setting, robustness has been investigated using Bayesian methods \citep{zhu2011robust}, trading the mean for the median \citep{cadre2001convergent}, using bounded loss functions \citep{maronna2013robust}, or leveraging principal component analysis \citep{kalogridis2019robust}.

In the operator-valued kernel literature, $\epsilon$-insensitive losses for vector-valued regression have been proposed by \citet{sangnier2017data} for finite-dimensional outputs.
The use of convex optimization tools such as the infimal convolution operator and parametric duality leads to efficient solvers and provides sparse estimators.
This idea is exploited by \citet{laforgue2020duality} where a generalization of this approach to infinite-dimensional outputs encompassing both the Huber and $\epsilon$-insensitive losses is developed.

In this paper, we extend the families of losses considered by \citet{laforgue2020duality} by leveraging specific $p$-norms in functional spaces to handle various forms of outliers (with Huber loss) and sparsity (with $\epsilon$-insensitive losses).
We study the properties of their Fenchel-Legendre conjugates, and derive the associated dual optimization problems, which require suitable representations and approximations adapted to each situation to be manageable computationally.
We propose tractable optimization algorithms for $p \in \{1,2\}$ in the Huber loss scenario, and for $p\in \{2, +\infty\}$ with the $\epsilon$-insensitive family.
Finally, we provide an empirical study of the proposed algorithms over synthetic and real functional datasets.

The paper is structured as follows. After introducing the general problem in \Cref{sec:background}, we focus in \Cref{sec:huber} on a generalized family of Huber losses and propose loss-specific tractable optimization schemes, before turning to the family of $\epsilon$-insensitive losses in \Cref{sec:epsilon}.
We illustrate the benefits of the approach on several benchmarks in \Cref{sec:experiments}.
Proofs are deferred to the supplement.

\section{Problem Formulation}
\label{sec:background}
In this section, we introduce the general setting of FOR in the context of vv-RKHSs, chosen for their modeling flexibility.
To benefit from duality principles, we focus on losses that can be expressed as infimal convolutions in the functional output space.
\paragraph{Notations:} Let $\mX$ be an input set,
$\Theta \subset \reals$  a compact set endowed with a Borel probability measure $\mu$, $\mY:=L^2[\Theta, \mu]$ the space of square $\mu$-integrable real-valued functions.
For $p \in [1, +\infty[$ and $f \in \mY$, let $\norm{f}_p = \left[ \int_\Theta |f(\theta)|^p \mathrm{d}\mu(\theta) \right]^\frac{1}{p} \in [0, +\infty]$; $\norm{\cdot}_\infty$ refers to the essential supremum.
In both cases, the norm is allowed to take infinite value ($\norm{\cdot}_p \colon \mY \to \R \cup \{+\infty\}$).\footnote{This assumption is natural in convex optimization which is designed to handle functions taking infinite values.}
Two numbers $p$ and $q \in [1, +\infty]$ are said to be conjugate exponents if $\frac{1}{p} + \frac{1}{q} = 1$, with the classical $0=\frac{1}{\infty}$ convention.
The ball in $\mY$ of radius $\epsilon > 0$ and center $0$ w.r.t.\ $\norm{\cdot}_p$ is denoted by $\cB^p_{\epsilon}$.
The space of bounded linear operators over $\mY$ is $\mL(\mY)$.
An operator-valued kernel (OVK) $K \colon \mX \times \mX \to \mL(\mY)$ is a mapping such that $\sum_{i=1}^n \sum_{j=1}^n \langle K(x_i, x_j) y_i, y_j \rangle_\mY \geq 0$ for all $(x_i, y_i)_{i=1}^n \subset \mX \times \mY$ and positive integer $n$.
An OVK $K$ gives rise to a space of functions from $\mX$ to $\mY$ called vector-valued RKHS (vv-RKHS); it is defined as $\mH_K := \overline{\Span}\{K(\cdot,x)y \, : \, (x,y) \in \mX \times \mY\}$, where $\Span(\cdot)$ denotes the linear hull of its argument, $\bar{\cdot}$ stands for closure, and $K(\cdot,x)y$ is the function $x' \in \mX \mapsto K(x',x)y\in  \mY$ while keeping $x\in \mX$ fixed.
The \emph{Fenchel-Legendre conjugate} of a function $f \colon \mY \to \left[-\infty, +\infty\right]$ is defined as $f^\star(z) := \sup_{y \in \mY} ~ \langle z, y \rangle_{\mY} - f(y)$ where $z \in \mY$.
Given a convex set $\cC \subset \mY$,
$\iota_\cC(\cdot)$ is its indicator function ($\iota_\cC(x)=0$ if $x\in \cC$, and $\iota_\cC(x) = + \infty$ otherwise), and $\Proj_\cC(\cdot)$ is the orthogonal projection on $\cC$ when $\cC$ is also closed.
Given two functions $f,g \colon \mY \to \left]-\infty, +\infty\right]$, their \emph{infimal convolution} is defined as $f \infconv g ~ (y) = \inf_{y' \in \mY} f(y - y') + g(y')$ for $y\in \mY$ and the \emph{proximal operator} of $f$ (when $f$ is convex, lower semi-continuous) is $\prox_f(y) = \argmin_{y' \in \mY} \frac{1}{2} \norm{y-y'}_\mY^2 + f(y')$ for all $y\in \mY$.
For a positive integer $n$, let $[n]=\{1,\ldots,n\}$.
For $p\in [1,\infty[$, the $p$-norm of a vector $\b v \in \R^m$ is denoted by $\norm{\b{v}}_p = \left( \sum_{j=1}^m |v_j|^p \right)^{\frac{1}{p}}$, and $\norm{\b{v}}_\infty = \max_{j \in [m]}{|v_j|}$.
Given a matrix $\b A \in \R^{n \times m}$ and $p,s \in [1, +\infty]$, $\norm{\b A}_{p,s}$ is the $s$-norm of the $p$-norms of the rows of $\b A$. The positive part of $x\in \R$ is denoted by $|x|_+=\max(x,0)$.

Next we introduce the FOR problem in vv-RKHSs.
Recall that $\mX$ is a set and $\mY=L^2[\Theta, \mu]$, the latter capturing the functional outputs.
Assume that we have \iid samples $(x_i, y_i)_{i\in[n]}$ from a random variable $(\rvX, \rvY) \in \mX \times \mY$.
Given a proper, convex lower-semicontinuous loss function $L \colon \mY \to \R$ and a regularization parameter $\lambda > 0$, we consider the \emph{regularized empirical risk minimization} problem
\begin{problem}
  \label{pbm:primal}
    \inf_{h \in \mH_K} ~ \frac{1}{n}\sum_{i \in [n]} L(y_i - h(x_i)) + \frac{\lambda}{2} \norm{h}_{\mH_K}^2,
\end{problem}
where $K \colon \mX \times \mX \to \mL(\mY)$ is a \emph{decomposable} OVK of the form $K = k_\mX T_{k_\Theta}$.
Here $k_\mX \colon \mX \times \mX \to \R$ and $k_\Theta \colon \Theta \times \Theta \to \R$ are continuous real-valued kernels, and $T_{k_\Theta} \in \mL(\mY)$ is the integral operator associated to $k_\Theta$, defined for all $f \in \mY$ as $(T_{k_\Theta} f)(\theta) := \int_\Theta k(\theta, \theta') f(\theta) \mathrm{d}\mu(\theta')$ where  $\theta \in \Theta$.
Similarly to the scalar case \citep{wahba90spline}, the minimizer of \Cref{pbm:primal} enjoys a representer theorem \citep{micchelli05learning} and writes as
\begin{equation*}
  \hat{h} = \frac{1}{\lambda n}\sum_{i \in [n]} k_\mX(\cdot, x_i) T_{k_\Theta}\hat{\alpha}_i
\end{equation*}
for some coefficients $\{\hat{\alpha}_i\}_{i \in [n]} \subset \mY$.
However, the functional nature of these parameters renders the problem extremely challenging, with quite few existing solutions.
Particularly, even in the case of the square loss
\begin{equation*}
  L(f) = \frac{1}{2}\norm{f}_\mY^2= \frac{1}{2} \int_\Theta f(\theta)^2 \mathrm{d}\mu(\theta),
\end{equation*}
the value of $\{\hat{\alpha}_i\}_{i\in [n]}$ can not be computed in closed form, and some level of approximation is required.
For instance in \citet{lian2007nonlinear}, $L$ is approximated as a discrete sum, allowing for a double application of the representer theorem and yielding tractable models.
In \citet{kadri2016ovk}, the integral operator is traded for a finite rank approximation based on its eigendecomposition, providing a computable closed-form expression for the coefficients.
Aiming at robustness, \citet{laforgue2020duality} propose a Huber loss based on infimal convolution, yet limited to a narrow choice of kernels $k_\Theta$.
Moreover, the lack of flexibility in the definition of the loss prevents the resulting estimators from being robust to a large variety of outliers.

The \tb{goal} of this  work is (i) to widen the scope of the FOR problem (\Cref{pbm:primal}) by considering losses capable of handling different forms of outliers and sparsity,
and (ii) to design efficient optimization schemes for the resulting tasks.
The proposed two loss families are based on infimal convolution and can be written in the form
\begin{align}
    \frac{1}{2} \norm{\cdot}_\mY^2 \infconv g \label{eq:loss-form}
\end{align}
for appropriately chosen functions $g \colon \mY \to ]-\infty, +\infty]$.
The key property which allows one to handle these convoluted losses from an optimization perspective is the fact that
\begin{equation*}
  \left(\frac{1}{2} \norm{\cdot}_\mY^2 \infconv g\right)^\star = \frac{1}{2} \norm{\cdot}_\mY^2 + g^\star,
\end{equation*}
as it makes the associated \Cref{pbm:primal} amenable for dual approaches.
The losses with the proposed dedicated optimization schemes are detailed in Section~\ref{sec:huber} and Section~\ref{sec:epsilon}, respectively.

The starting point for working with convoluted losses in vv-RKHSs is the following lemma.
\begin{lemma}[Dualization for convoluted losses;  \citealt{laforgue2020duality}]
  \label{lemma:dual_pbm}
 Let $L$ be a loss function defined as $L = \frac{1}{2} \norm{\cdot}_\mY^2 \infconv g$ for some $g \colon \mY \to ]-\infty, +\infty]$.
The solution of \Cref{pbm:primal} is given by
\begin{equation}\label{eq:dual_estimator}
\hat{h} = \frac{1}{\lambda n}\sum_{i\in [n]} k_\mX(\cdot, x_i) T_{k_\Theta}\hat{\alpha}_i,
\end{equation}
with $(\hat{\alpha}_i)_{i\in [n]} \in \mY^n$ being the solution of the dual task
\begin{problem}\label{pbm:dual}
\begin{aligned}
\inf_{(\alpha_i)_{i\in [n]}\in \mY^n} & \sum_{i\in [n]} \left[ \frac{1}{2} \norm{\alpha_i}^2_\mY - \langle \alpha_i, y_i \rangle_\mY +  g^\star(\alpha_i) \right]  \\
&  + \frac{1}{2\lambda n} \sum_{i,j \in [n]}  k_\mX(x_i, x_j) \left\langle \alpha_i, T_{k_\Theta} \alpha_j\right\rangle_\mY.
\end{aligned}
\end{problem}
\end{lemma}
Solving \Cref{pbm:dual} in the general case for various $g$ and $k_\Theta$ raises multiple \tb{challenges} which have to be handled simultaneously.
\Cref{pbm:dual} is often referred to as a \emph{composite} optimization problem, with a differentiable term consisting of a quadratic part
added to a non-differentiable term induced by $g^\star$.
The first challenge is to be able to compute the proximal operator associated to $g^\star$.
The second and third difficulties arise from the fact that the dual variables are functions ($\alpha_i \in \mY$) and hence managing them computationally requires specific care.
Particularly, evaluation of $\left\langle \alpha_i, T_{k_\Theta} \alpha_j\right\rangle_\mY$ can be non-trivial.
In addition, one has to design a finite-dimensional description of the dual variables that is compatible with $T_{k_\Theta}$ and the proximal operator of $g^\star$.
The primary focus and technical contribution of the paper is to handle these challenges, after which proximal gradient descent optimization can be applied.
We detail our proposed solution in the next section.

\section{Learning with $\mY$-Huber Losses}
\label{sec:huber}
In this section, we propose a generalized Huber loss
on $\mY$ based on infimal convolution, followed by an efficient dual optimization approach to solve the corresponding \Cref{pbm:primal}.
This loss (as illustrated in Section~\ref{sec:experiments}) shows robustness against different kind of outliers.
Our proposed loss on $\mY$ relies on functional $p$-norms where $p \in [1, +\infty]$.
\begin{definition}[$\mY$-Huber loss]
  \label{def:huber}
  Let $\kappa > 0$ and $p \in [1, +\infty]$.
  We define the Huber loss with parameters $(\kappa, p)$ as
  \begin{equation*}
    H_{\kappa}^p := \frac{1}{2} \norm{\cdot}_\mY^2 \infconv \kappa \norm{\cdot}_p.
  \end{equation*}
\end{definition}
Notice that in the specific case of $\mY=\R$, $H_{\kappa}^p$ reduces to the classical Huber loss on the real line for arbitrary $p$.
Our following result describes the behavior of $H_{\kappa}^p$.
\begin{proposition}
  \label{prop:huber_explicit}
  Let $\kappa > 0$, $p \in [1, +\infty]$, and $q$ the conjugate exponent of $p$.
  Then for all $f \in \mY$,
  \begin{equation*}
    H_{\kappa}^p(f) = \frac{1}{2} \norm{\Proj_{\cB_\kappa^q}(f)}_\mY^2 + \kappa \norm{f - \Proj_{\cB_{\kappa}^q}(f)}_p.
  \end{equation*}
\end{proposition}
\tb{Remark:}
 For general $p$, the value of $H_{\kappa}^p(f)$ can not be computed straighforwardly due to the complexity of the projection on $\cB_{\kappa}^q$.
 As we show however using a dual approach \Cref{pbm:primal} is still computationally manageable.
 For $p=2$, one gets back the loss investigated by \citet{laforgue2020duality}.

The following proposition is a key result of this work that allows to leverage $p$-norms as suitable candidates for $g$ in \eqref{eq:loss-form}.
It extends to $\mY$ the well-known finite-dimensional case that can \eg be found in 
\citet{bauschke2011convex}.
\begin{proposition}
  \label{prop:fl_p_norm}
  Let $p,q \in [1, +\infty]$ such that $\frac{1}{p} + \frac{1}{q} = 1$.
  Then
  \begin{equation*}
    \norm{\cdot}_p ^\star = \iota_{\cB^q_1}(\cdot).
  \end{equation*}
\end{proposition}
Our next result provides the dual of \Cref{pbm:primal}, and shows the impact of the parameters $(p, \kappa)$.
\begin{proposition}[Dual Huber]
  \label{prop:dual_huber}
  Let $\kappa > 0$, $p \in [1, +\infty]$, and $\frac{1}{p}+\frac{1}{q}=1$.
  The dual of \Cref{pbm:primal} with loss $H_\kappa^p$ writes
  \begin{problem}
    \label{pbm:dual_huber}
    \begin{aligned}
      \inf_{(\alpha_i)_{i=1}^n \in \mY^n}
      &\sum_{i=1}^n \left[\frac{1}{2}\norm{\alpha_i}_{\mY}^2 - \langle \alpha_i, y_i \rangle_\mY\right] \\
      &  + \frac{1}{2 \lambda n} \sum_{i=1}^n \sum_{j=1}^n k_{\mX}(x_i, x_j) \left \langle \alpha_i, T_{k_\Theta} \alpha_j \right \rangle_\mY  \\
      & \mathrm{s.t} \, \norm{\alpha_i}_q \leq \kappa, \quad i \in [n].
    \end{aligned}
  \end{problem}
\end{proposition}
\tb{Remarks:}
\begin{itemize}[labelindent=0em,leftmargin=*,topsep=0cm,partopsep=0cm,parsep=0cm,itemsep=0mm]
    \item Influence of $\kappa$ and $p$: The difference between using the square loss and the Huber loss $H_\kappa^p$ lies in the constraint on the $q$-norm of the dual variables.
    The parameter $p$ influences the \emph{shape} of the ball via the dual exponent $q$ ($\norm{\alpha_i}_q$) defining the admissible region for dual variables, and $\kappa$ determines its \emph{size} ($\norm{\alpha_i}_q \leq \kappa$).
    As $\kappa$ grows, the constraint becomes void and we recover the solution of the classical ridge regression problem.
    In \Cref{sec:loss_illustration}, we explore how different choices of $p$ can affect the sensitivity of the loss to two different types of outliers.
    \item Partially observed data: The observed data $(y_i)_{i\in[n]}$ enter into \Cref{pbm:dual_huber} only via their scalar product with the dual variables $(\alpha_i)_{i\in [n]}$.
    However in real life scenarii one never fully observe the $y_i$ functions and these inner products are to be estimated.
    One can instead assume access to a sampling at some locations $(\theta_j)_{j=1}^m$ which can be used to approximate the inner products (see \Cref{sec:huber_discrete}).
\end{itemize}
Let us now recall the \tb{challenges} to be tackled to solve \Cref{pbm:dual_huber}.
Firstly, as $\mY$ is infinite-dimensional no finite parameterization of the dual variables can be assumed a priori.
Secondly, even computing the different terms of the objective function is non-trivial.
Indeed, computing the quadratic term corresponding to the regularization is not straightforward as it involves the terms $\left \langle \alpha_i, T_{k_\Theta} \alpha_j \right \rangle_\mY$, which require the knowledge of the action of the integral operator $T_{k_\Theta}$.
The third difficulty comes from handling the constraints.
Gradient-based optimization algorithms will require the projection of the dual variables on the feasible set $\cB_{\kappa}^q \subset \mY$, which can be intractable to evaluate for some choices of $q$ depending on the chosen representation.

The next proposition ensures the tractability of the projection step for specific choices of $q$.
\begin{proposition}[Projection on $\cB_{\kappa}^q$]
  \label{prop:projection_q_ball}
  Let $\kappa > 0$.
  The projection on $\cB_{\kappa}^q$ is tractable for $q=2$ and $q=\infty$ and can be expressed for all $(\alpha, \theta) \in \mY \times \Theta$ as
  \begin{align}
    \label{eq:proj_norm_2}
    \Proj_{\cB_\kappa^2}(\alpha) &= \min{\left(1, \frac{\kappa}{\norm{\alpha}_2}\right)} \alpha,\\
    \label{eq:proj_norm_inf}
    \left(\Proj_{ \cB_\kappa^\infty }(\alpha) \right)(\theta) &= \sign{(\alpha(\theta))} \min{(\kappa, |\alpha(\theta)|)}.
  \end{align}
\end{proposition}
The projection operator for $p=q=2$ in \Cref{eq:proj_norm_2} simply consists of a multiplication by a scalar involving the $2$-norm of the dual variable.
In the $p=1$ case (i.e., $q=\infty$), the projection \Cref{eq:proj_norm_inf} involves a \emph{pointwise} projection.
A suitable representation must guarantee the feasibility of this projection, which requires a pointwise control over the dual variables.

In order to solve \Cref{pbm:dual_huber}, we propose to use two different representations.
In \Cref{sec:huber_discrete}, we advocate representing the dual variables by linear splines and approximating the action of $T_{k_\Theta}$ by Monte-Carlo (MC) sampling.
Splines allow pointwise control of the dual variable which is well-suited to both $p=1$ and $p=2$.
Our alternative approach (elaborated in Section~\ref{sec:huber_eigen}) relies on a finite-rank approximation of $T_{k_\Theta}$ using its eigendecomposition.
This method is applicable when $p=2$ with the complementary advantage of performing dimensionality reduction.
\begin{table*}[t]
  \centering
    \caption{Correspondence between the quantities involved in \Cref{pbm:dual} depending on the representation.}
        \begin{tabular}{c|c|c}
           & Linear splines & Eigenvectors of $T_{k_\Theta}$\\
          \hline
          $\sum_{i  \in [n]} \frac{1}{2} \norm{\alpha_i}_\mY^2$ & $ \frac{1}{2m}\Tr \left( \b{A} \b{A}^\top \right)$
          & $\Tr \left( \frac{1}{2} \b{A} \b{A}^\top \right)$
           \vspace{0.07cm} \\
          $\sum_{i\in [n]} \langle \alpha_i, y_i \rangle_\mY$ & $\frac{1}{m}\Tr \left( \b{A} \b{Y}^\top \right)$ &$\Tr \left( \b{A} \b{R}^\top \right)$
          \vspace{0.1cm} \\
          $ \sum_{i,j \in [n]} k_{\mX}(x_i, x_j) \left \langle \alpha_i, T_{k_\Theta} \alpha_j \right \rangle_\mY$ &
          $\frac{1}{m^2}\Tr \left(  \b{K}_\mX \b{A} \b{K}_\Theta \b{A}^\top \right)$ &$\Tr \left(  \b{K}_\mX \b{A} \b{\Delta} \b{A}^\top \right)$
        \end{tabular}
  \label{tab:correspondency_optim}
\end{table*}
\subsection{The Linear Spline Based Approach}
\label{sec:huber_discrete}
In this section we introduce a linear spline based representation for the dual variables to tackle the challenges outlined.

Linear splines represent an easy-to-handle function class which provides pointwise control over the dual variables as they are encoded by their evaluations at some knots.
While the class lacks smoothness, the dual variables are smoothed out by $T_{k_\Theta}$ in the estimator expression from \Cref{eq:dual_estimator}.
Indeed, given that the kernel $k_\Theta$ is $2s$-times continuously differentiable, the RKHS $\mH_{k_\Theta}$ (where $T_{k_\Theta}$ maps) consists of $s$-times continuously differentiable functions \citep{zhou08derivative}, a desirable property in many settings making linear splines good candidates for modeling the dual variables.

A linear spline is a piecewise linear curve which can be encoded by a set of ordered locations or anchor points $(\theta_j)_{j \in [m]} \in \Theta^m$, and by a vector of size $m$ corresponding to the evaluation of the spline at these points.
We choose the anchors to be distributed \iid according to $\mu$; in practice, we often take the locations to be those of the available sampling of the observed data $(y_i)_{i \in[n]} \in \mY^n$.

Fixing the anchors, the $n$ dual variables are encoded by the matrix of evaluations $\b{A} = [\b{a}_i]_{i\in [n]} =[\alpha_i(\theta_j)]_{i\in [n], j\in [m]}\in \R^{n \times m}$ with $\b a_i$ being the $i^{th}$ row of $\b A$.
The action of $T_{k_\Theta}$ on a function $\alpha \in \mY$ is then approximated using MC approximation as
\begin{equation*}
  T_{k_\Theta} \alpha \approx \frac{1}{m} \sum_{j \in [m]} k_\Theta(\cdot, \theta_j) \alpha(\theta_j),
\end{equation*}
resulting in the estimator
\begin{align*}
  h(x)(\theta) = \frac{1}{\lambda n m} \sum_{i\in [n]} k_\mX(x, x_i) \sum_{j\in [m]} a_{ij}  k_\Theta(\theta, \theta_j).
\end{align*}
Using this $\b{A}$ parameterization of $h$, the different terms in \Cref{pbm:dual_huber} are approximated as follows.
\begin{itemize}[labelindent=0em,leftmargin=*,topsep=0cm,partopsep=0cm,parsep=0cm,itemsep=0mm]
    \item \tb{Squared norm of the dual variables}:
    We approximate the squared norm of the dual variables using MC sampling with locations $(\theta_j)_{j\in [m]}$:
    \begin{equation*}
      \sum_{i\in [n]} \frac{1}{2} \norm{\alpha_i}_\mY^2 \approx \frac{1}{2m} \Tr \left( \b{A}^\top \b{A} \right).
    \end{equation*}
    \item \tb{Scalar product with the data:}
        We encode the evaluations of the observed functions $(y_i)_{i\in [n]}$ at locations $(\theta_j)_{j\in [m]}$ in a matrix $\b{Y} = [y_i(\theta_j)]_{i \in [n],\, j\in [m]} \in \R^{n \times m}$, and again use MC approximation
        \begin{equation*}
          \sum_{i\in [n]} \langle \alpha_i, y_i \rangle_\mY \approx \frac{1}{m} \Tr \left ( \b{A} \b{Y}^\top \right).
        \end{equation*}
    \item \tb{Regularization term:}
    Encoding the dual variables as linear splines hinders the exact computation of the quadratic terms $\langle \alpha_i, T_{k_\Theta} \alpha_j \rangle_\mY$, which we propose to approximate using a MC approximation of $T_{k_\Theta}$.
    Letting $\b{K}_\mX \in \R^{n \times n}, \b{K}_\Theta \in \R^{m \times m}$ be the Gram matrices respectively associated to the data $(x_i)_{i \in [n]}$ and kernel $k_\mX$, and to the locations $(\theta_j)_{j\in [m]}$ and kernel $k_\Theta$, the regularization term can be rephrased as $\frac{1}{\lambda n m^2}\Tr \left(  \b{K}_\mX \b{A} \b{K}_\Theta \b{A}^\top \right)$.
    \item \tb{Constraints:}
    The constraints $\norm{\alpha_i}_q \leq \kappa$ can be handled similarly. Particularly, let $\alpha_i \in \mS_m$ with associated evaluations vector $\b{a}_i \in \R^m$ and $q \in [1, +\infty[$. We trade the integral expression in the norm for an MC sum, resulting in the constraint $\norm{\b{a}_i}_q \leq m^{\frac{1}{q}} \kappa$.
    This expression also holds for $q=+\infty$, as $\norm{\alpha_i}_{\infty} \leq \kappa$ iff.\ $\norm{\b{a}_i}_{\infty} \leq \kappa$.
\end{itemize}
Gathering the different terms (summarized in Table~\ref{tab:correspondency_optim}) yields the following relaxation of  \Cref{pbm:dual_huber}:
\begin{problem}
  \label{pbm:dual_huber_splines}
  \begin{aligned}
    \inf_{\b{A} \in \R^{n \times m}} &\Tr \left (
    \frac{1}{2} \b{A} \b{A}^\top
    - \b{A} \b{Y}^\top
    + \frac{1}{2\lambda n m} \b{K}_\mX \b{A} \b{K}_\Theta \b{A}^\top \right )
    \\&\st \norm{\b{A}}_{q, \infty} \leq m^{\frac{1}{q}} \kappa.
  \end{aligned}
\end{problem}
\tb{Remark:} The decomposable assumption on the kernel $K$ plays a role in the regularization. It has the effect of disentangling the action of both Gram matrices $\b{K}_\mX$ and $\b{K}_\Theta$.

We propose to tackle \Cref{pbm:dual_huber_splines} using accelerated proximal gradient descent (APGD), where the proximal step amounts to projecting the coefficients on the $q$-ball of radius $m^{1/q}\kappa$.
The technique is summarized in \Cref{alg:pgd_splines}.
The gradient stepsize $\gamma$ can be computed exactly from the parameters of the problem.
Indeed, for guaranteed convergence, one must set $\gamma < \frac{2}{C}$ where $C$ is the Lipschitz constant associated to the gradient of the objective function; here $C \leq 1 + \frac{1}{\lambda n} \norm{\b{K_{\mX}}}_{\mathrm{op}} \norm{\b{K_{\Theta}}}_{\mathrm{op}}.$
The initialization can either be the null matrix $\b{A}^{(0)} =  \b 0_{\R^{n \times m}}$ or the solution of the unconstrained optimization problem which can be obtained in closed form.
This solution (i) can dramatically reduce the number of iterations for small $\epsilon$ or large $\kappa$, and (ii) can be computed in $\mathcal O (n^3 + m^3)$ time exploiting the Kronecker structure inherited from the separable kernel with a Sylvester equation solver \citep{Sima96, DinuzzoAl11}.
Since the objective function in \Cref{pbm:dual_huber_splines} is the sum of two functions, one convex and differentiable with Lipschitz continuous gradient (the quadratic form) and one convex and lower semi-continuous (the indicator function of the constraint set), the optimal worst case complexity is $\mathcal O \left ( \frac{1}{T^2} \right )$ \citep{teboulleFista09}.
The time complexity per iteration is dominated by the computation of the matrix $\b{K}_\mX \b{V} \b{K}_\Theta$ which is $\mathcal O(n^2m + m^2 n)$.
\LinesNumberedHidden{
\begin{algorithm}[t]
\SetKwInOut{Input}{input}
\SetKwInOut{Init}{init}
\SetKwInOut{Parameter}{Param}
\SetInd{0.5em}{0.5em}
\caption{APGD with linear splines
\label{alg:pgd_splines}}
\Input{~Gram matrices $\b{K}_\mX, \b{K}_\Theta$, data matrix $\b{Y}$, regularization parameter $\lambda$, loss parameters $(\kappa, p)$ or $(\epsilon, p)$, gradient step $\gamma$}
\Init{$\b{A}^{(0)}, \b{A}^{(-1)} = \bm{0} \in \mathbb{R}^{n \times m}$}
\For{epoch $t$ from $0$ to $T-1$}{
    \tcp{gradient step}
    $\b{V} = \b{A}^{(t)} + \frac{t - 2}{t + 1} \left (\b{A}^{(t)} - \b{A}^{(t-1)}  \right)$

    $\b{U} = \b{V} - \gamma\left( \b{V} + \frac{1}{\lambda n m}\b{K}_\mX \b{V} \b{K}_\Theta - \b{Y} \right)$

    \tcp{projection step}
    \If{$p=2$}{
    \For{row $i \in [n]$}{
    $\b{a}_{i}^{(t+1)} = \min\left(\frac{\sqrt{m}\kappa}{\norm{\b{u}_{i}}_2}, 1\right) \b{u}_{i}$
    \tcp{if $H_\kappa^2$
    }
    $\b{a}_{i}^{(t+1)} =
  \left | 1 - \frac{\gamma \epsilon}{\sqrt{m}\norm{\b{u}_{i}}_2} \right |_+ \b{u}_i$
    \tcp{if $\ell_\epsilon^2$}
    }
    }
    \Else{
    \For{row $i \in [n]$}{
      \For{column $j \in [m]$}{
        $a_{ij}^{(t+1)} = \sign (u_{ij} ) \min\left(\kappa, \left|u_{ij}\right|\right)$
        \tcp{if $H_\kappa^1$}
        $a_{ij}^{(t+1)} = \sign{(u_{ij})} \left | |u_{ij}| - \frac{\gamma \epsilon}{m} \right |_+$
        \tcp{if $\ell_\epsilon^\infty$}
        }
      }
    }
    }
\Return{$\b{A}^{(T)}$}
\end{algorithm}}
%
\subsection{The Eigendecomposition Approach}
\label{sec:huber_eigen}
In this section, we propose an alternative finite-dimensional description of the dual variables relying on an approximate eigendecomposition of $T_{k_\Theta}$ when $p=2$.
The rationale of this approach is to decrease the number of parameters needed to represent the estimator by selecting directions well-suited to $T_{k_\Theta}$, namely the dominant $r$ eigenvectors of $T_{k_\Theta}$.
As computing the eigendecomposition of $T_{k_\Theta}$ is generally intractable, we propose an approximation detailed in the following.

Let us consider the problem of finding a continuous eigenvector $\psi$ of a sampled version of the integral operator $T_{k_\Theta}$ with eigenvalue $\delta > 0$ \citep{hoegaerts2005subset}:
\begin{equation*}
  \frac{1}{m} \sum_{j\in [m]} k_\Theta(\theta, \theta_j) \psi(\theta_j) = \delta \psi(\theta) \text{ for }\forall \theta \in \Theta.
\end{equation*}
By evaluating it at points $(\theta_j)_{j\in [m]}$, one gets that $(\delta m, \psi(\theta_j)_{j\in [m]})$ form the eigensystem of the Gram matrix $\b{K}_\Theta \in \R^{m \times m}$.
These can be computed using for instance singular value decomposition, and by substitution one arrives at an approximated eigenbasis of dimension at most $m$; this can be used as a proxy instead of the true eigenvectors of $T_{k_\Theta}$.
By using the first $r$ of these vectors for some $r < m$ we are able to lower the size of the parameterization of the model.
We store in a diagonal matrix $\bm{\Delta} \in \R^{r \times r}$ the first $r$ eigenvalues.

The problem is now parameterized by a matrix $\b{A} = [\b{a}_i]_{i \in [n]} \in \R^{n \times r}$ with each row $\b{a}_i \in \R^r$ encoding the coefficients of the dual variable $\alpha_i$ on the $(\psi_j)_{j\in [r]}$ basis.
The estimator then reads as
\begin{equation*}
  h(x)(\theta) = \frac{1}{\lambda n} \sum_{i\in [n],\, j\in [r]}  a_{ij} \delta_j k_\mX(x, x_i) \psi_j(\theta).
\end{equation*}
We store in $\b{R} = [R_{ij}]_{i\in [n], j\in [r]}\in \R^{n \times r}$ the scalar products between the observed data and the eigenbasis: $R_{ij} = \langle y_i, \psi_j \rangle_\mY$.
The correspondence between the different optimization terms are summarized in \Cref{tab:correspondency_optim}; the optimization reduces to
\begin{align}
  \label{pbm:dual_huber_eigen}
    \inf_{\b{A} \in \R^{n \times r}} ~~&
    \Tr \left( \frac{1}{2} \b{A} \b{A}^\top - \b{A} \b{R}^\top
    + \frac{1}{2 \lambda n} \b{K}_\mX \b{A} \b{\Delta} \b{A}^\top \right) \nonumber \\
    & \st \norm{\b{A}}_{2,\infty} \leq \kappa.
\end{align}
Again, one can use APGD to solve this task; the resulting computations are
deferred to \Cref{alg:pgd_eigen} in the supplement.

\tb{Remark:}
For $\kappa$ large enough, the solution reduces to that of \citet{kadri2016ovk}.
However, our approximation allows handling a wide range of kernels in contrast to the analytical knowledge imposed for the eigensystem of $T_{k_\Theta}$ by \citet{kadri2016ovk,laforgue2020duality}.

\section{Learning with $\epsilon$-insensitive Losses}
\label{sec:epsilon}
In this section, we propose a generalized $\epsilon$-insensitive version of the square loss on $\mY$ involving infimal convolution, and derive tractable dual optimization algorithm to solve \Cref{pbm:primal}.
This loss induces sparsity on the matrix of coefficients as illustrated in \Cref{sec:experiments}.
\begin{definition}[$\epsilon$-insensitive loss]
  \label{def:epsilon}
  Let $\epsilon > 0$ and $p \in [1, +\infty]$.
  We define the $\epsilon$-insensitive version of the square loss with parameters $(\epsilon, p)$ as
  \begin{equation*}
    \ell_{\epsilon}^p := \frac{1}{2} \norm{\cdot}_\mY^2 \infconv \iota_{\cB^p_{\epsilon}}(\cdot).
  \end{equation*}
\end{definition}
When $\mY=\R$, the loss reduces to the classical $\epsilon$-insensitive version of the square loss regardless of $p$.
The following proposition (counterpart of \Cref{prop:huber_explicit}) sheds light on the effect of the infimal convolution on the square loss.
\begin{proposition} \label{prop:l_eps_p}
  Let $\epsilon > 0$ and $p \in [1, +\infty]$. Then for all $f \in \mY$,
  \begin{equation}
  \ell_\epsilon^p(f) = \frac{1}{2}\norm{f -\Proj_{\cB_\epsilon^p}(f)}^2_\mY.
\end{equation}
\end{proposition}
\tb{Remark:}
 Proposition~\ref{prop:l_eps_p} means that $\ell_\epsilon^p(f) = 0$ when $\norm{f}_p \leq \epsilon$, i.e.\ small residuals do not contribute to the risk.
For general $p$, $\ell_\epsilon^p(f)$ is not straightforward to compute due to the complexity of $\Proj_{\cB_\epsilon^p}(f)$.
As we however use a dual approach, \Cref{pbm:primal} can still be tackled computationally.

The next result shows how to dualize \Cref{pbm:primal} when the proposed $\epsilon$-insensitive loss is used.
\begin{proposition}[Dual $\epsilon$-insensitive]
  \label{prop:dual_eps}
  Let $\epsilon \geq 0, p \in [1, +\infty]$, and $\frac{1}{p}+\frac{1}{q}=1$. The dual of \Cref{pbm:primal} writes as
  \begin{problem}
    \label{pbm:dual_epsilon}
    \begin{aligned}
      \inf_{(\alpha_i)_{i\in [n]} \in \mY^n} &
      \sum_{i \in [n]} \left[\frac{1}{2}\norm{\alpha_i}_{\mY}^2 - \langle \alpha_i, y_i \rangle_\mY + \epsilon \norm{\alpha_i}_q\right] \\
      & + \frac{1}{2 \lambda n} \sum_{i\in [n],\, j\in [n]} k_{\mX}(x_i, x_j) \left \langle \alpha_i, T_{k_\Theta} \alpha_j \right \rangle_\mY. \nonumber
    \end{aligned}
  \end{problem}
\end{proposition}
\tb{Remark} (influence of $\epsilon$ and $p$):
  Compared to the square loss, $\ell_\epsilon^p$ induces an additional term $\sum_{i\in [n]} \epsilon \norm{\alpha_i}_q$ in the dual.
  Setting $\epsilon=0$ recovers the square loss case.

The challenges involving the representation of the dual variables, and the computability of the different terms composing \Cref{pbm:dual_epsilon} are similar to those evoked in \Cref{sec:huber}.
We have however traded the constraints on the $q$-norms of the dual variables against an additional non-smooth term.
As for the Huber loss family in \Cref{sec:huber}, we address this convex non-smooth optimization problem through the APGD algorithm.
The proximal step involves the computation of $\prox_{\gamma \epsilon \norm{\cdot}_q}$ for a suitable gradient stepsize $\gamma > 0$, which is the focus of the next proposition.
\begin{proposition}[Proximal $q$-norm]
  \label{prop:proximal_q_norm}
Let $\epsilon > 0$. The proximal operator of $\epsilon \norm{\cdot}_q$ is computable for $q=1$ and $q=2$, and given for all $(\alpha, \theta) \in \mY \times \Theta$ by
\begin{align}
  \label{eq:prox_norm_1}
  \left(\prox_{ \epsilon \norm{\cdot}_1 }(\alpha) \right)(\theta) &= \sign{(\alpha(\theta))} \left| \left|\alpha(\theta)\right| - \epsilon \right|_+, \\
  \label{eq:prox_norm_2}
  \prox_{\epsilon \norm{\cdot}_2 }(\alpha) &= \alpha
  \left | 1 - \frac{ \epsilon}{\norm{\alpha}_\mY} \right |_+.
\end{align}
\end{proposition}
We recognize in \Cref{eq:prox_norm_1} a pointwise soft thresholding, and in \Cref{eq:prox_norm_2} an analogous to the block soft thresholding, both are known to promote sparsity.

To solve \Cref{pbm:dual_epsilon} we rely on the two kinds of finite representations introduced previously.
In \Cref{sec:epsilon_discrete}, we tackle the $p=2$ and $p=\infty$ case with the linear splines based method from \Cref{sec:huber_discrete}, before using dimensionality reduction from \Cref{sec:huber_eigen} for the case $p=2$ in \Cref{sec:epsilon_eigen}.
\subsection{The Linear Spline Based Approach}
\label{sec:epsilon_discrete}
Similarly to what was presented in \Cref{sec:huber_discrete}, we use linear splines to represent the dual variables $(\alpha_i)_{i\in[n]}$ as they allow a pointwise control over the dual variable and thus give rise to computable proximal operators.
Keeping the notations, the optimization boils down to
\begin{align*}
  \label{pbm:dual_epsilon_splines}
    \inf_{\b{A} \in \R^{n \times m}} &\Tr \left (
    \frac{1}{2} \b{A} \b{A}^\top
    - \b{A} \b{Y}^\top
    + \frac{1}{2\lambda n m} \b{K}_\mX \b{A} \b{K}_\Theta \b{A}^\top \right ) \\
    & + \frac{\epsilon}{m^{\frac{1}{q}}} \sum_{i\in [n]} \norm{\b{a}_i}_q.
\end{align*}
We use APGD to solve it with steps detailed in \Cref{alg:pgd_splines}.
When $q=1$, the proximal operator is the soft thresholding operator, akin to promote sparsity in the dual coefficients.
\subsection{The Eigendecomposition Approach}
\label{sec:epsilon_eigen}
We mobilize the eigendecomposition technique from \Cref{sec:huber_eigen} to solve \Cref{pbm:dual_epsilon} in the case $p=2$.
Using the same notation as in Problem \ref{pbm:dual_huber_eigen}, we get the following task:
\begin{equation*}
  \label{pbm:dual_epsilon_eigen}
  \begin{aligned}
    \hspace{-0.1cm}\inf_{\b{A} \in \R^{n \times r}} \hspace{-0.2cm}
    \Tr \left( \frac{1}{2} \b{A} \b{A}^\top \hspace{-0.2cm} - \b{A} \b{R}^\top \hspace{-0.2cm}
    + \frac{1}{2 \lambda n} \b{K}_\mX \b{A} \b{\Delta} \b{A}^\top \hspace{-0.1cm}\right) \hspace{-0.1cm} + \hspace{-0.05cm}\epsilon \sum_{i\in [n]} \norm{\b{a}_i}_2.
  \end{aligned}
\end{equation*}
APGD is applied to tackle this problem; the details are deferred to \Cref{alg:pgd_eigen} in the supplement.
Notice that the proximal operator in this case is the block soft thresholding operator, known to promote structured row-wise sparsity.

\section{Numerical Experiments}
\label{sec:experiments}
In this section, we demonstrate the efficiency of the proposed convoluted losses.
The implementation is done in Python, and is available in the form of an open source package at \href{https://github.com/allambert/foreg}{https://github.com/allambert/foreg}.\\
The experiments are centered around \tb{two key directions}:
\begin{enumerate}[labelindent=0em,leftmargin=*,topsep=0cm,partopsep=0cm,parsep=0cm,itemsep=0mm]
  \item The first goal is to understand the  accuracy-sparsity tradeoffs of the $\epsilon$-insensitive loss as a function of the regularization $\lambda$ and insensitivity parameter $\epsilon$.
  \item Our second aim is to quantify the robustness of the Huber losses w.r.t.\  different forms of outliers with a particular focus on global versus local ones.
  To gain further insight into the  robustness w.r.t.\ corruption, we designed 3 types of functional outliers with distinct characteristics.
\end{enumerate}
Our proposed losses are investigated on 3 benchmarks: a synthetic one associated to Gaussian processes, followed by two real-world ones arising in the context of neuroimaging and speech analysis.
We investigate both questions on the synthetic data, and provide further insights for the first and the second question on the neuroimaging and the speech dataset, respectively.

We now detail the 3 functional outlier types used in our experiments on robustness to study the effect of local and global corruption.
Local outliers affect the functions only on small portions of $\Theta$ whereas global ones contaminates them in their entirety.
To corrupt the functions $(y_i)_{i\in [n]}$, we first draw a set $I \subset \{0,\ldots,n\}$ of size $\lfloor \tau n \rfloor$ corresponding to the indices to contaminate; $\tau \in [0, 1]$ being the proportion of contaminated functions.
Then, we perform different kinds of corruption:\\
$\bullet$ \textbf{Type 1}: Let $\omega$ be the permutation defined for $j \in [|I|]$ as $\omega(I_j) = I_{j+1}$ if $j < |I|$ and $\omega(I_{|I|}) = I_1$, then for $i \in I$, the data point $(x_i, y_i)$ is replaced by $(x_i, - y_{\omega(i)})$.\\
$\bullet$ \textbf{Type 2}: Given covariance parameters $\pmb \sigma \in \mathbb R^r$  and an intensity parameter $\zeta > 0$, we draw a Gaussian process $g_c \sim \mathcal{GP} (0, k_{\sigma_c})$ for $c \in [r]$ where $k_{\sigma_c}$ is the Gaussian covariance function with standard deviation $\sigma_c$.
Then, for $i \in I$, we replace $(x_i, y_i)$ with $\left(x_i, \sum_{c\in [r]} a_{ic} g_c\right) $ where the coefficients $a_{ic}$ are drawn i.i.d.\ from a uniform distribution $\mathcal U([-0.5 \zeta, 0.5 \zeta])$.\\
$\bullet$ \textbf{Type 3}: For each $i \in I$, a randomly chosen fraction $\xi \in [0, 1]$ of the discrete observations for $y_i$ is replaced by random draws from a uniform distribution $\mathcal U([-b_{\text{max}}, b_{\text{max}}])$, where $b_{\text{max}}:=\underset{i, j}{\max} |y_i(\theta_j)|$.
The corruptions of Type 1 and 2 are global ones whereas Type 3 is a local one.
In terms of the characteristics of the different corruptions, for Type 1 the properties of the outlier functions remain close to those of the non-outlier ones, whereas with Type 2 they become completely different.
Finally, for corrupted data in the hyperparameter choice using cross-validation the mean was replaced with median.

For the losses $H_\kappa^1$ and $\ell_\epsilon^\infty$ we solve the problem based on the representation with linear splines (see \Cref{sec:huber_discrete} and \Cref{sec:epsilon_discrete} respectively); this is the only possible approach.
However for the losses $H_\kappa^2$ and $\ell_\epsilon^2$ we exploit the representation using a truncated basis of approximate eigenfunctions (see \Cref{sec:huber_eigen} and \Cref{sec:epsilon_eigen} respectively), in doing so we reduce the computational cost compared to the linear splines approach.
Concerning optimization, we deployed the APGD method \citep{teboulleFista09} with backtracking line search, and adaptive restart \citep{candes15restart}. The initialization in APGD was carried out with the closed-form solution available for the square loss using a Sylvester solver.
\begin{table*}
  \centering
\caption{MSEs and sparsity on the DTI dataset} \label{tab:dti}
\begin{small}
\begin{sc}
\setlength\tabcolsep{1.7pt}
\begin{tabular}{lllllll}
\toprule
    $\lambda$ & Metric & $\nicefrac{1}{2} \lVert \cdot \rVert_\mY^2$ & $H_\kappa^2 $ & $H_\kappa^1$ & $\ell^2_\epsilon$ & $\ell_\epsilon^\infty$ \\
    \midrule
   10$^{-5}$	& MSE (10$^{-1}$) & 2.5$\pm$0.19 & \textbf{2.21}$\pm$0.31 & \textbf{2.21}$\pm$0.31 & 2.41$\pm$0.26 & 2.5$\pm$0.23 \\
   	& Sparsity & - & - & - & 27.4$\pm$17.2\% & 85.9$\pm$10.7\% \\
   	\midrule
   10$^{-3}$ & MSE (10$^{-1}$) & \textbf{2.18}$\pm$0.27 & 2.23$\pm$0.32 & 2.21$\pm$0.32 & 2.2$\pm$0.29 & \textbf{2.18}$\pm$0.28 \\
   	& Sparsity & - & - & - & 3.4$\pm$6.9\% & 12.7$\pm$10.5\% \\
\end{tabular}
\end{sc}
\end{small}
\end{table*}

Regarding the performance measure applied for evaluation, let $((y_i(\theta_{ij}))_{j\in[m_i]})_{i\in [n]}$ be the set of observed discretized functions and let $(\widehat{y}_i(\theta_{ij}))_{j \in [m_i]})_{i\in [n]}$ be an estimated set of discretized functions, where  $(\theta_{ij})_{j\in [m_i]}$ denotes the observation locations for $y_i$.
We used the mean squared error defined as
\begin{equation*}
  \text{MSE}:=\frac 1 n \sum_{i \in [n]} \sum_{j \in [m_i]} [y_i(\theta_{ij}) - \widehat{y}_i(\theta_{ij})]^2.
\end{equation*}
When $m_i = m$ for all $i$, we normalize it by $m$ and define NMSE$:=\frac 1 m$ MSE.
\subsection{Experiments on the synthetic dataset}
\label{subsec:toy}
The impact of the different losses are investigated in detail on a function-to-function synthetic dataset whose construction is detailed in \Cref{sec:synth}.
The kernels $k_{\mX}$ and $k_\Theta$ are chosen to be Gaussian and the experiments are averaged over 20 draws with
training and testing samples of size 100.
\begin{figure}
\begin{center}
\includegraphics[width=1\linewidth]{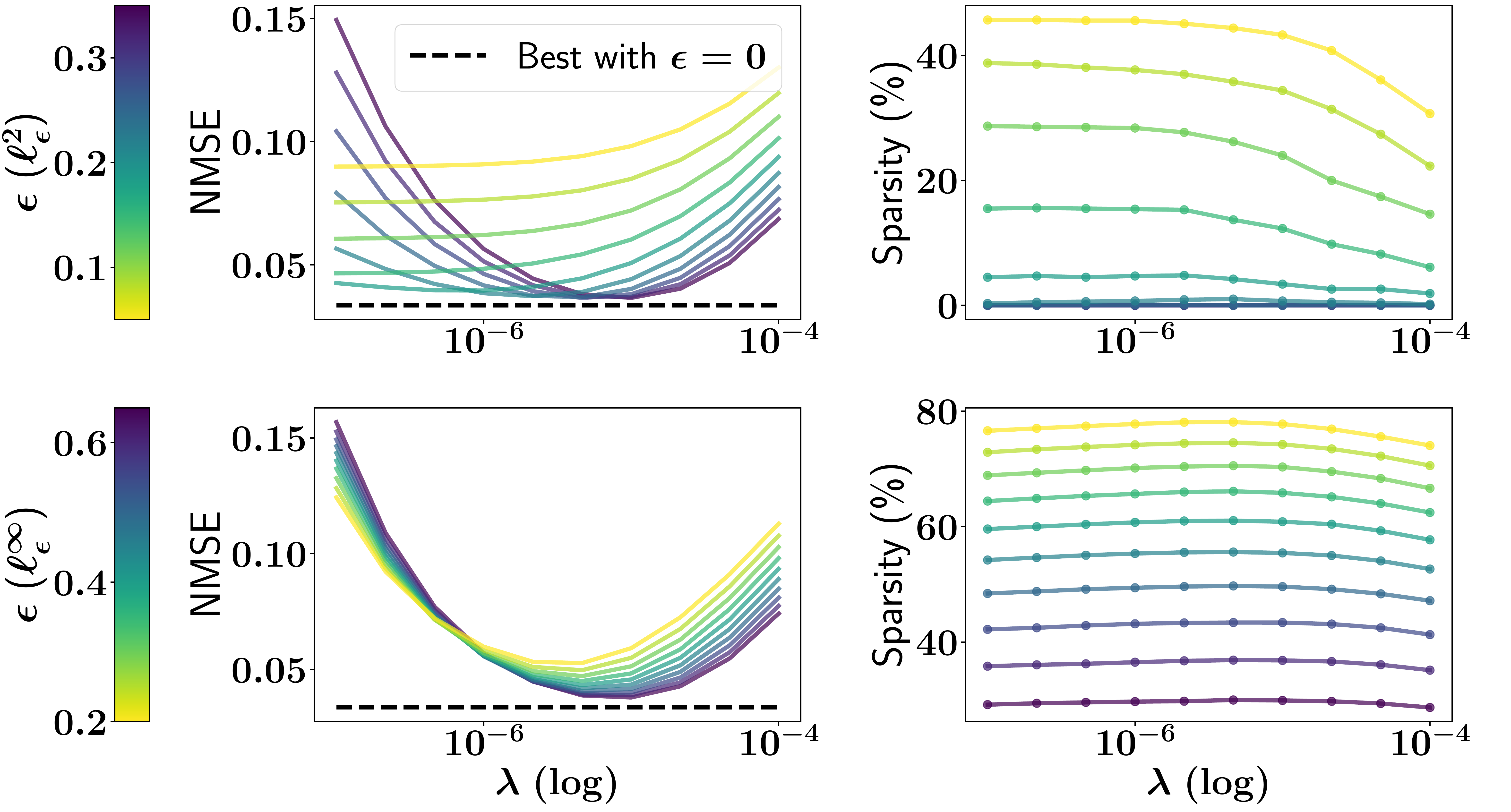}
\end{center}
\caption{Interaction between regularization $\lambda$ and insensitivity $\epsilon$ for the loss $\ell_\epsilon^2$ (1st row) and $\ell_\epsilon^\infty$ (2nd row).}
\label{fig:toyeps}
\end{figure}

\tb{$\epsilon$-insensitive loss:} To study the interaction between $\lambda$ and $\epsilon$ and the resulting sparsity-accuracy trade-offs, we added i.i.d.\ Gaussian noise with standard deviation $0.5$ to the observations of the output functions.
The resulting MSE values are summarized in Fig.~\ref{fig:toyeps}.
For both the $\ell_\epsilon^2$ and the $\ell_\epsilon^\infty$ loss, one can reduce $\lambda$, increase $\epsilon$ and get a fair amount of sparsity while making a small compromise in terms of accuracy.

\begin{figure}
\begin{center}
\includegraphics[width=1\linewidth]{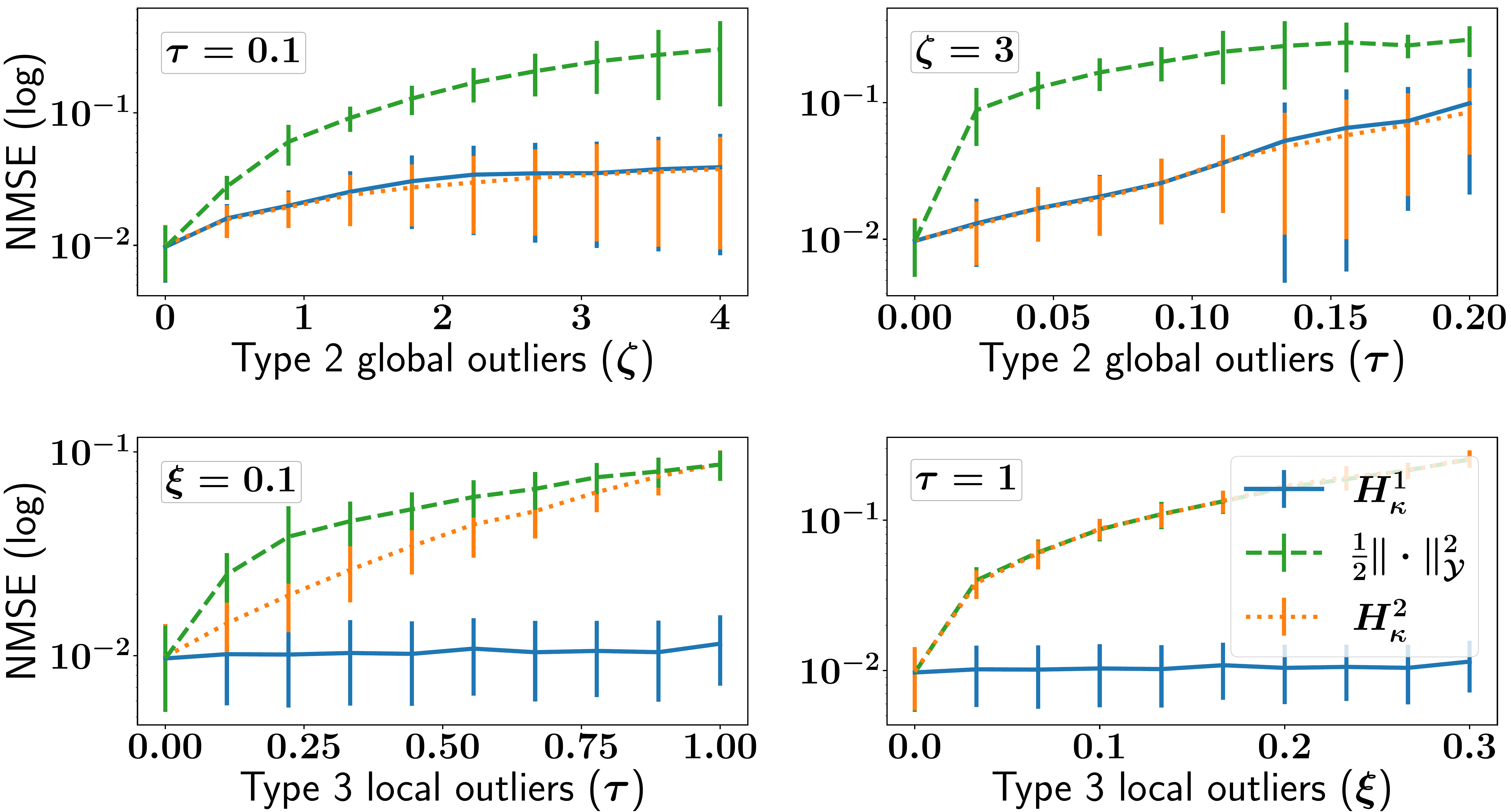}
\end{center}
\caption{Robustness to different types of outliers.}
\label{fig:toyout}
\end{figure}

\tb{Huber loss}: We investigate the robustness of the Huber loss to different types of outliers while selecting both $\lambda$ and $\kappa$ through robust cross validation.
The resulting MSE values are summarized in Fig.~\ref{fig:toyout}.
As it can be seen in the first row of the figure, the losses $H_\kappa^1$ and $H_\kappa^2$ are significantly more robust to global outliers than the square loss, both when the outliers' intensity $\zeta$ and the proportion $\tau$ of contaminated samples vary.
The second row of the figure shows that when dealing with local outliers, the closer one gets to the whole sample being contaminated ($\tau=1$), the less robust $H_\kappa^2$ becomes.
On the other hand, $H_\kappa^1$ shows remarkable robustness as it can be observed at the bottom right panel (in case of $\tau = 1$).
One can interpret this phenomenon by noticing that the loss $H_\kappa^2$ can penalize less big discrepancy between functions in the $\|. \|_{\mY}$ norm sense, but if all samples are contaminated locally a little, the outliers are meddled in the norm and so $H_\kappa^2$ becomes inefficient.
%
\subsection{Experiments on the DTI dataset}
\begin{table*}[t]
  \centering
    \caption{MSEs on speech data} \label{tab:speech}
\begin{small}
\begin{sc}
\setlength\tabcolsep{2.5pt}
\begin{tabular}{llllll}
\toprule
    VT & $\nicefrac{1}{2} \lVert \cdot \rVert_\mY^2$ & $H_\kappa^2 $ & $H_\kappa^1$ & $\ell^2_\epsilon$ & $\ell_\epsilon^\infty$ \\
    \midrule
    LP & \textbf{6.58}$\pm$0.62 & 6.59$\pm$0.62 & 6.59$\pm$0.64 & \textbf{6.58}$\pm$0.62 & \textbf{6.58}$\pm$0.62 \\ 
    LA & 4.65$\pm$0.55 & 4.65$\pm$0.55 & 4.66$\pm$0.55 & \textbf{4.64}$\pm$0.55 & \textbf{4.64}$\pm$0.55 \\ 
    TBCL & \textbf{4.26}$\pm$0.46 & \textbf{4.26}$\pm$0.46 & 4.27$\pm$0.46 & \textbf{4.26}$\pm$0.46 & \textbf{4.26}$\pm$0.46 \\ 
    TBCD & 4.67$\pm$0.37 & 4.68$\pm$0.38 & 4.7$\pm$0.38 & \textbf{4.67}$\pm$0.38 & \textbf{4.67}$\pm$0.38 \\ 
    VEL & \textbf{2.94}$\pm$0.5 & \textbf{2.94}$\pm$0.5 & 2.95$\pm$0.5 & \textbf{2.94}$\pm$0.5 & \textbf{2.94}$\pm$0.5 \\ 
    GLO & \textbf{7.25}$\pm$0.65 & 7.26$\pm$0.65 & \textbf{7.25}$\pm$0.64 & \textbf{7.25}$\pm$0.65 & \textbf{7.25}$\pm$0.65 \\ 
    TTCL & 3.76$\pm$0.21 & 3.76$\pm$0.21 & 3.74$\pm$0.2 & \textbf{3.73}$\pm$0.21 & \textbf{3.73}$\pm$0.21 \\ 
     TTCD & 5.93$\pm$0.34 & 5.94$\pm$0.34 & 5.93$\pm$0.35 & \textbf{5.92}$\pm$0.34 & \textbf{5.92}$\pm$0.34 \\ 
\end{tabular}
\end{sc}
\end{small}
\end{table*}
\begin{table*}
  \centering
\caption{MSEs on contaminated speech data} \label{tab:speechcorrupt}
\begin{small}
\begin{sc}
\setlength\tabcolsep{2.5pt}
\begin{tabular}{lllllll}
\toprule
    \multirow{2}{*}{VT} & \multicolumn{3}{c}{Type 1 Outliers ($\tau=0.1$)} & \multicolumn{3}{c}{Type 3 outliers ($\tau=0.1$, $\xi=0.1$)} \\
    \cmidrule(lr){2-7}
    & $\nicefrac{1}{2} \lVert \cdot \rVert_\mY^2$ & $H_\kappa^2 $ & $H_\kappa^1$ & $\nicefrac{1}{2} \lVert \cdot \rVert_\mY^2$ & $H_\kappa^2 $ & $H_\kappa^1$  \\
    \midrule
    LP & 9.4$\pm$0.75 & 9.4$\pm$0.66 & \textbf{9.19}$\pm$0.79 & 7.53$\pm$0.58 & 7.62$\pm$0.59 & \textbf{7.0}$\pm$ 0.59 \\
    LA & 5.72$\pm$0.76 & 5.63$\pm$0.71 & \textbf{5.52}$\pm$0.69 & \textbf{5.06}$\pm$0.6 & 5.11$\pm$0.6 & 5.09$\pm$0.55 \\
    TBCL & 6.71$\pm$0.96 & 6.14$\pm$0.97 & \textbf{5.98}$\pm$0.93 & 5.06$\pm$0.51 & 5.16$\pm$0.48 & \textbf{4.72}$\pm$ 0.54 \\
    TBCD & \textbf{5.8}$\pm$0.41 & 5.86$\pm$0.44 & 5.83$\pm$0.44 & 5.18$\pm$0.4 & 5.26$\pm$0.41 & \textbf{5.08}$\pm$ 0.4 \\
    VEL & 4.37$\pm$0.56 & 3.76$\pm$0.62 & \textbf{3.76}$\pm$0.59 & 3.52$\pm$0.57 & 3.54$\pm$0.58 & \textbf{3.41}$\pm$ 0.57 \\
    GLO & 9.61$\pm$0.87 & \textbf{9.51}$\pm$0.86 & 9.53$\pm$0.84 & 7.94$\pm$0.61 & 8.02$\pm$0.61 & \textbf{7.76}$\pm$ 0.61 \\
    TTCL & 15.06$\pm$2.22 & 9.51$\pm$0.63 & \textbf{9.48}$\pm$0.6 & \textbf{5.89}$\pm$0.43 & 5.91$\pm$0.45 & 6.62$\pm$0.66 \\
     TTCD & 8.15$\pm$0.48 & \textbf{7.96}$\pm$0.49 & 8.02$\pm$0.51 & 6.63$\pm$0.44 & 6.74$\pm$0.42 & \textbf{6.36}$\pm$ 0.39 \\
\end{tabular}
\end{sc}
\end{small}
\end{table*}
\label{subsec:dti}
In our next experiment we considered the DTI benchmark\footnote{This dataset was collected at the Johns Hopkins University
and the Kennedy-Krieger Institute.}.
The dataset contains a collection of fractional anisotropy profiles deduced from diffusion tensor imaging scans, and we take the first scans of the $n=100$ multiple sclerosis patients.
The profiles are given along two tracts, the corpus callosum and the right cortiospinal.
The goal is to predict the latter function from the former, which can be framed as a function-to-function regression problem.
When some functions admit missing observations, we fill in the gaps by linear interpolation, and later use the MSE as metric.
We use a Gaussian kernel for $k_{\mX}$ and a Laplacian one for $k_\Theta$ and average over 10 runs with $n_{\text{train}}=70$ and $n_{\text{test}}=30$.

Similarly to our experiences gained on the synthetic dataset, a compromise can be made between the two parameters $\lambda$ and $\epsilon$ to get increased sparsity, as can be observed in Table \ref{tab:dti}.
Moreover, we highlight that even for optimal regularization with respect to the square loss $\lambda=10^{-3}$, one gets a fair amount of sparsity while getting the same score with $\ell_{\epsilon}^\infty$ and a very small difference with $\ell_{\epsilon}^2$.
\subsection{Speech data}
%
%
In this section, we focus on a speech inversion problem \citep{MitraAl09}. Particularly, our goal is to predict a vocal tract (VT) configuration that likely produced a speech signal \citep{Richmond02}.
This benchmark encompasses $n=413$ synthetically pronounced words to which 8 VT functions are associated: LA, LP, TTCD, TTCL, TBCD, TBCL, VEL, GLO. This is then a time-series--to--function regression problem. We predict the VT functions separately in eight subproblems.

Since the words are of varying length, we use the MSE as metric and extend symmetrically the signals to match the longest word for in training.
We encode the input sounds through 13 mel-frequency cepstral coefficients (MFCC) and normalize the VT functions' values to the range [-1, 1].
We average over $10$ train-test splits taking $n_{\text{train}}=250$ and $n_{\text{test}}=163$.
Finally we take an integral Gaussian kernel on the standardized MFCCs (see \Cref{sec:expes_supplement} for further details) as $k_\mX$ and a Laplace kernel as $k_\Theta$.

We first compare all the losses on untainted data in Table~\ref{tab:speech}.
Then to evaluate the robustness of the Huber losses, we ran experiments on contaminated data with two configurations.
In the first case, we added Type 1 (global) outliers with $\tau=0.1$ and in the second one, we added Type 3 (local) outliers with $\tau=0.1$ and $\xi=0.1$.
The results are displayed in Table~\ref{tab:speechcorrupt}.
In the contaminated setting, one gets results similar to ones obtained on the synthetic dataset.
The loss $H_\kappa^1$ works especially well for local outliers whereas the loss $H_\kappa^2$ is robust only to global outliers.


\section{Conclusion}
\label{sec:conclusion}
In this paper we introduced generalized families of loss functions based on infimal convolution and p-norms for functional output regression.
The resulting optimization problems were handled using duality principles.
Future work could focus on extending these techniques to a wider choice of $p$ using iterative techniques for the proximal steps.
\section*{Acknowledgements}
AL, DB, and FdB received funding from the \emph{Télécom Paris Research and Testing Chair on Data Science and Artificial Intelligence for Digitalized Industry and Service}. AL also obtained additional funding from the \emph{ERC Advanced Grant E-DUALITY (787960)}.


\bibliography{references}
\bibliographystyle{icml2022}

\newpage
\appendix
\onecolumn
The supplement is structured as follows.
We present the proofs of our results in \Cref{sec:proofs}.
\Cref{sec:expes_supplement} complements the main part of the paper by providing additional algorithmic and experimental details.
Finally, \Cref{sec:loss_plots} includes additional plots and insights about the loss functions.
\section{Proofs}
\label{sec:proofs}
In this section, we present the proofs of our results. In \Cref{sec:convex} we recall some definitions from convex optimization used throughout the proofs, with focus on Fenchel-Legendre conjugation and proximal operators, followed by the proofs themselves (\Cref{sec:proof_fl}-\ref{sec:proof6}).
\subsection{Reminder on Convex Optimization}
\label{sec:convex}
Recall that $\mY := L^2[\Theta, \mu]$ where $\Theta \subset \R$ is a compact set endowed with a probability measure $\mu$.
\begin{definition}[Proper, convex, lower semi-continuous functions]
We denote by $\Gamma_0(\mY)$ the set of functions $J \colon \mY \to \left]-\infty, + \infty\right]$ that are
\begin{enumerate}[labelindent=0em,leftmargin=*,topsep=0cm,partopsep=0cm,parsep=0cm,itemsep=0mm]
  \item \emph{proper}: $\dom(J) := \condset{f \in \mY}{J(f) < +\infty} \neq \varnothing$,
  \item \emph{convex}: $J(t f + (1 - t)g) \leq t J(f) + (1 - t) J(g)$ for $\forall f, g \in \mY, \forall t \in [0, 1]$, and
  \item \emph{lower semicontinuous}: $\varliminf_{g \to f} J(g) \geq J(f)$ for $\forall f \in \mY$, where $\varliminf$ denotes  limit inferior.
\end{enumerate}
\end{definition}
\begin{definition}
  \label{def:fl}
  The \emph{Fenchel-Legendre conjugate} of a function $J \colon \mY \to \left[-\infty, +\infty\right]$ is defined as
  \begin{align*}
    \label{eq:fl}
    J^\star(f)& := \sup_{g \in \mY} ~ \langle f, g \rangle_{\mY} - J(g), \quad f \in \mY.
  \end{align*}
\end{definition}
The Fenchel-Legendre conjugate of a function $J$ is always convex. It is also involutive on $\Gamma_0(\mY)$, meaning that $(J^\star)^\star = J$ for any $J \in \Gamma_0(\mY)$.
We gather in \Cref{tab:fl_useful} examples and properties of Fenchel-Legendre conjugates.

We now introduce the infimal convolution operator following \citet{bauschke2011convex}.
\begin{definition}[Infimal convolution]
  \label{def:infconv}
   The \emph{infimal convolution} of two functions $L,J \colon \mY \to \left]-\infty, +\infty\right]$ is
  \begin{equation*}
    L \infconv J \colon
    \left(
    \begin{aligned}
      \mY &\to [- \infty, + \infty] \\
       f  &\mapsto \inf_{g \in \mY} L(f - g) + J(g)
    \end{aligned}
    \right).
  \end{equation*}
\end{definition}
One key property of the infimal convolution operator is that it behaves nicely under Fenchel-Legendre conjugation, as it is  detailed in the following proposition.
\begin{lemma}[\citealt{bauschke2011convex}, Proposition~13.24] \label{prop:fl_infconv}
  Let $L,J \colon \mY \to \left]-\infty, +\infty\right]$. Then
  \begin{align*}
    (L \infconv J)^\star = L^\star + J^\star.
  \end{align*}
\end{lemma}
We now define the proximal operator, used as a replacement for the classical gradient step in the presence of non-differentiable objective functions.
\begin{definition}[Proximal operator, \citealt{moreau65prox}] The \emph{proximal operator} (or proximal map) is defined as
\begin{align}
  \label{eq:prox}
  \prox_J (f) &:= \argmin_{g \in \mY} ~ J(g) + \frac{1}{2} \norm{f-g}^2_\mY, \text{ for } (J,f) \in \Gamma_0(\mY) \times \mY.
\end{align}
\end{definition}
One advantage of working with functions in $\Gamma_0(\mY)$ is that the proximal operator is always well-defined.
Its computation is doable for various losses thanks to the following lemma.
\begin{lemma}[Moreau decomposition, \citealt{moreau65prox}]
\label{lemma:moreau}
Let $J \in \Gamma_0(\mY)$ and $\gamma > 0$. Then
\begin{equation}
  \label{eq:moreau}
    \Id = \prox_{\gamma J}(\cdot) + \gamma \prox_{J^\star/\gamma}(\cdot/\gamma),
\end{equation}
where $\Id$ stands for the identity operator.
\end{lemma}
\begin{table}[t]
  \centering
    \caption{Properties of Fenchel-Legendre conjugate, for any $J,L \colon \mY \to \left[-\infty, +\infty\right]$ and $p,q \in [1, +\infty]$ s.t.\ $\frac{1}{p} + \frac{1}{q} = 1$.  \label{tab:fl_useful}}
  \begin{tabular}{cc}
    \toprule
       Function & Fenchel-Legendre conjugate \\
    \midrule
     $\frac{1}{2} \norm{\cdot}_{\mY}^2$ & $\frac{1}{2} \norm{\cdot}_{\mY}^2$ \\
     \addlinespace
     $\norm{\cdot}_p$ & $\iota_{\cB^{q}_1}$\\
     \addlinespace
     $\epsilon J$ & $\epsilon J^\star(\frac{\cdot}{\epsilon})$ for all $\epsilon > 0$ \\
     \addlinespace
     $J(\cdot - y)$ & $J^\star + \langle \cdot, y \rangle_{\mY}$ for all $y \in \mY$\\
     \addlinespace
     $L \infconv J$ & $L^\star + J^\star$\\
    \bottomrule
  \end{tabular}
\end{table}
We remind the reader that we want to solve
\begin{problem}
  \label{pbm:primal_sup}
    \inf_{h \in \mH_K} ~ \frac{1}{n}\sum_{i\in [n]} L(y_i - h(x_i)) + \frac{\lambda}{2} \norm{h}_{\mH_K}^2,
\end{problem}
where $K \colon \mX \times \mX \to \mL(\mY)$ is a \emph{decomposable} OVK of the form $K = k_\mX T_{k_\Theta}$.
Here $k_\mX \colon \mX \times \mX \to \R$ and $k_\Theta \colon \Theta \times \Theta \to \R$ are continuous real-valued kernels, and $T_{k_\Theta} \in \mL(\mY)$ is the integral operator associated to $k_\Theta$, defined for all $\alpha \in \mY$ by
\begin{equation*}
   (T_{k_\Theta} \alpha)(\theta) = \int_\Theta k(\theta, \theta') \alpha(\theta) \mathrm{d}\mu(\theta')\text{ for all } \theta \in \Theta.
\end{equation*}
We also remind the reader to the dual of \Cref{pbm:primal_sup} when the loss writes as an infimal convolution.
\begin{lemma}[Dualization for convoluted losses, \citealt{laforgue2020duality}]
  \label{lemma:dual_pbm_sup}
 Let $L$ be a loss function defined as $L = \frac{1}{2} \norm{\cdot}_\mY^2 \infconv g$ for some $g \colon \mY \to ]-\infty, +\infty]$.
Then the solution of \Cref{pbm:primal_sup} is given by
\begin{equation}\label{eq:dual_estimator_sup}
\hat{h} = \frac{1}{\lambda n}\sum_{i\in [n]} k_\mX(\cdot, x_i) T_{k_\Theta}\hat{\alpha}_i,
\end{equation}
with $(\hat{\alpha}_i)_{i\in [n]} \in \mY^n$ being the solution of the dual task
\begin{problem}\label{pbm:dual_sup}
\begin{aligned}
\inf_{(\alpha_i)_{i\in [n]}\in \mY^n} ~ \sum_{i\in [n]}  \left[\frac{1}{2} \norm{\alpha_i}^2_\mY - \langle \alpha_i, y_i \rangle_\mY + g^\star(\alpha_i)\right] + \frac{1}{2\lambda n} \sum_{i,j \in [n]}  k_\mX(x_i, x_j) \left\langle \alpha_i, T_{k_\Theta} \alpha_j\right\rangle_\mY.
\end{aligned}
\end{problem}
\end{lemma}
\subsection{Proof of \Cref{prop:fl_p_norm}}
\label{sec:proof_fl}
Before going through the proof, let us recall  Hölder's inequality.
\begin{lemma}[Hölder's inequality]
Let $p, q \in [1, +\infty]$ be conjugate exponents, in other words $\frac{1}{p} + \frac{1}{q} = 1$. Let $\Theta$ be a measurable space enriched with probability measure $\mu$. Then for any $f, g: \Theta \rightarrow \R$ measurable functions one has
\begin{equation*}
  \int_\Theta \abs{ f(\theta) g(\theta)} \mathrm{d}\mu(\theta) \leq \norm{f}_p \norm{g}_q.
\end{equation*}
Moreover, if $p \in ]1, +\infty[$, $f \in L^p[\Theta, \mu]$ and $g \in L^q[\Theta, \mu]$, then equality is attained if and only if $\abs{f}^p$ and $\abs{g}^p$ are linearly dependent in $L^1[\Theta, \mu]$.
\end{lemma}
We now introduce a lemma useful to the proof of \Cref{prop:fl_p_norm}.
\begin{lemma}
  \label{lemma:intermediate_fl}
  Let $p, q \in ]1, +\infty[$ be conjugate exponents and $f \in \mY$ such that $1 < \norm{f}_q < +\infty$.
  Then there exist $h \in \mY$ and $C > 0$ such that
  \begin{equation*}
    \langle f, h \rangle_\mY - \norm{h}_p \geq C.
  \end{equation*}
  Moreover, one can choose $h$ such that whenever $f(\theta)=0$, $h(\theta)=0$ also holds.
\end{lemma}
\paragraph{Proof of \Cref{lemma:intermediate_fl}}
Let $p, q \in ]1, +\infty[$ be conjugate exponents and $f \in \mY$ such that $1 < \norm{f}_q < +\infty$.
We know that Hölder's inequality becomes an equality if and only if $\abs{f}^q$ and $\abs{g}^p$ are linearly dependent in $L^1[\Theta, \mu]$.
To that end, let $g \colon \Theta \to \R$ be defined as
\begin{align}
 g(\theta) = \sign(f(\theta)) \abs{f(\theta)}^{\frac{q}{p}} \quad \text{where $\theta \in \Theta$}. \label{eq:g-def}
\end{align}
It is to be noted that $g$ does not necessarily belong to $\mY$, yet it belongs to $L^p[\Theta, \mu]$.
By construction, we have
\begin{equation}
    \int_\Theta f(\theta)g(\theta) \mathrm{d}\mu(\theta) = \norm{f}_q \norm{g}_p.
    \label{eq:Holder-eq}
\end{equation}
We consider a sequence $(g_n)_{n \in \bbN} \in \mY^\bbN$ such that $g_n(\theta) = \sign{(g(\theta))} \min{(\abs{g(\theta)}, n)}$ with $(n, \theta) \in \bbN \times \Theta$.
As $\abs{g_n(\theta)} \leq n$ for all $(n, \theta) \in \bbN \times \Theta$ and $\mu$ is a probability measure, the functions $g_n$ belong to $\mY$.
Since (i) $g_n(\theta) \xrightarrow{n \to \infty} g(\theta)$ for all $\theta \in \Theta$ and (ii) $\abs{g_n(\theta)} \leq \abs{g(\theta)}$ for any $n \in \bbN$ holds $\mu$-almost everywhere, the dominated convergence theorem in $L^p[\Theta, \mu]$ ensures that $\norm{g - g_n}_p \xrightarrow{n \to \infty}0$.
Consequently, it holds that for all $n \in \bbN$,
\begin{align*}
  \abs{ \int_\Theta f(\theta) g(\theta) \mathrm{d}\mu(\theta) - \int_\Theta f(\theta) g_n(\theta) \mathrm{d}\mu(\theta)} &=
  \abs{ \int_\Theta f(\theta) [g(\theta) - g_n(\theta)] \mathrm{d}\mu(\theta)}
  \stackrel{(a)}{\leq} \int_\Theta \abs{f(\theta)} \abs{g(\theta) - g_n(\theta)} \mathrm{d}\mu(\theta)\\
  & \stackrel{(b)}{\leq} \norm{f}_q \norm{g - g_n}_p.
\end{align*}
In (a) we used that the absolute value of the integral can be upper bounded by the integral of the absolute value, in (b) the Hölder's inequality was invoked.
Thus by $\norm{g - g_n}_p \xrightarrow{n \to \infty}0$ and $\norm{f}_q < +\infty$, this means that $\langle f, g_n \rangle_\mY \xrightarrow{n \to \infty}  \int_\Theta f(\theta)g(\theta) \mathrm{d}\mu(\theta) \stackrel{\eqref{eq:Holder-eq}}{=} \norm{f}_q \norm{g}_p $, and for all $\epsilon > 0$, there exist $N \in \bbN$ such that for all $n \geq N$, $\langle f, g_n \rangle_\mY > (\norm{f}_q - \epsilon) \norm{g}_p$.
In particular for $\epsilon = \frac{\norm{f}_q - 1}{2}>0$, we have $\langle f, g_N \rangle_\mY > \frac{1 + \norm{f}_q}{2} \norm{g}_p$. Then,
\begin{align*}
  \langle f, g_N \rangle_\mY - \norm{g_N}_p &\stackrel{(c)}{\geq} \langle f, g_N \rangle_\mY - \norm{g}_p \stackrel{(d)}{\geq} \frac{1 + \norm{f}_q}{2} \norm{g}_p - \norm{g}_p \geq \underbrace{\frac{\norm{f}_q - 1}{2} \norm{g}_p}_{> 0}.
\end{align*}
In (c) we used that $\norm{g_N}_p \leq \norm{g}_p$, (d) is implied by $\langle f, g_N \rangle_\mY > \frac{1 + \norm{f}_q}{2} \norm{g}_p$. Taking $h = g_N$ and $C = \frac{\norm{f}_q - 1}{2} \norm{g}_p$ yields the announced result, by noticing that \eqref{eq:g-def} shows that $f(\theta) = 0$ also implies $h(\theta)=g_N(\theta) = g(\theta)=0$.
\qed

We are now ready to prove \Cref{prop:fl_p_norm}, which is the building block for dualizing optimization problems resulting from the generalized Huber and $\epsilon$-insensitive losses whose definition can be found respectively in \Cref{def:huber} and \Cref{def:epsilon}.
The proposition is an extension of the well-studied finite-dimensional case to the space $\mY$.
\begin{proposition*}[\ref{prop:fl_p_norm}]
  Let $p,q \in [1, +\infty]$ such that $\frac{1}{p} + \frac{1}{q} = 1$. Then
  \begin{equation}
    \label{eq:fl_tranform_sup}
    \norm{\cdot}_p ^\star = \iota_{\cB^q_1}(\cdot).
  \end{equation}
\end{proposition*}
\paragraph{Proof}
The proof is structured as follows.
We first consider the case of $p=1$, followed by $p\in ]1,+\infty]$, and $p=+\infty$.
The reasoning in all cases rely heavily on Hölder's inequality.
Throughout the proof it is assumed that $f \in \mY$.
\paragraph{Case $p=1$:}
The reasoning goes as follows: we show that  $\norm{f}_\infty \leq 1$ implies $\norm{\cdot}_1^\star (f) = 0$, and $\norm{f}_\infty > 1$ gives $\norm{\cdot}_1^\star (f) = +\infty$, which allow one to conclude that $\norm{\cdot}_1 ^\star = \iota_{\cB^\infty_1}(\cdot)$.
\begin{itemize}
  \item \tb{When $\norm{f}_\infty \leq 1$}: Exploiting Hölder's inequality, it holds that
  \begin{equation*}
    \langle f, g \rangle_\mY \leq \norm{f}_\infty \norm{g}_1 \quad \text{ for all } g \in \mY.
  \end{equation*}
  Since $\norm{f}_\infty \leq 1$, this implies that
  \begin{equation*}
    \langle f, g \rangle_\mY - \norm{g}_1 \leq 0 \quad \text{ for all } g \in \mY.
  \end{equation*}
  The supremum being attained for $g=0$, we conclude that $\norm{\cdot}_1^\star(f) = 0$.
  \item \tb{When $\norm{f}_\infty > 1$}: Let $A = \left\{ \theta \in \Theta\, : \, \abs{f(\theta)} > \frac{1 + \norm{f}_\infty}{2} \right\}$. By the definition of the essential supremum, $\mu(A) > 0$.
  We define $g \colon \Theta \to \R$ to be the function: $g(\theta)=\sign{(f(\theta))}$ if $\theta \in A$ and $0$ otherwise.
  Since $g$ is bounded, $g \in \mY$.
  Denoting by $t>0$ a running parameter, it holds that
  \begin{align*}
    \langle f, tg \rangle_\mY - \norm{tg}_1 & \stackrel{(a)}{=} \langle f, tg \rangle_\mY - t \mu(A) = t\int_{\Theta} f(\theta) g(\theta) \d \mu(\theta) - t \mu(A) \stackrel{(a)}{=}  t\int_A \abs{f(\theta)} \d \mu(\theta)- t \mu(A)\\ &    \stackrel{(b)}{\geq} t \mu(A) \frac{1 + \norm{f}_\infty}{2} - t \mu(A)
    = t  \underbrace{\mu(A) \frac{\norm{f}_\infty - 1}{2}}_{>0} \xrightarrow{t \to \infty} +\infty.
  \end{align*}
  In (a) we used the definition of $g$, (b) is implied by the fact that $\abs{f(\theta)} > \frac{1 + \norm{f}_\infty}{2}$ for all $\theta \in A$.
  Thus $\norm{\cdot}_1^\star(f) = +\infty$, which concludes the proof.
\end{itemize}
\paragraph{Case $p \in ]1, +\infty[$:} The reasoning proceeds as follows: we show that (i) $\norm{f}_q \leq 1$ implies $\norm{\cdot}_p^\star (f) = 0$, (ii) $1 < \norm{f}_q < +\infty$ gives $\norm{\cdot}_p^\star (f) = +\infty$, and (iii) $\norm{f}_q = +\infty$ results in $\norm{\cdot}_p^\star (f) = +\infty$.
This allows us to conclude that $\norm{\cdot}_p^\star = \iota_{\cB_1^q}(\cdot)$.
\begin{itemize}
\item \tb{When $\norm{f}_q \leq 1$}: By Hölder's inequality, it holds that
\begin{equation*}
  \langle f, g \rangle_\mY \leq \norm{f}_q \norm{g}_p \text{ for all } g \in \mY.
\end{equation*}
Exploiting $\norm{f}_q \leq 1$, we get that
\begin{equation*}
  \langle f, g \rangle_\mY -  \norm{g}_p \leq 0 \text{ for all }  g \in \mY.
\end{equation*}
The supremum being reached for $g=0$; we conclude that $\norm{\cdot}_p ^\star (f) = 0$.

\item \tb{When $1 < \norm{f}_q < +\infty$}: According to \Cref{lemma:intermediate_fl}, there exist $g \in \mY$ and $C > 0$ such that
\begin{equation*}
  \langle f, g \rangle_\mY - \norm{g}_p \geq C.
\end{equation*}
Denoting by $t>0$ a running parameter, one arrives at
\begin{equation*}
  \langle f, tg \rangle_\mY - \norm{tg}_p \geq tC \xrightarrow{t \to \infty} +\infty.
\end{equation*}
This shows that $\norm{\cdot}_p ^\star (f) = +\infty$.
\item \tb{When $\norm{f}_q = +\infty$}: We consider the sequence of functions $(f_n)_{n \in \bbN}$ defined as  $f_n(\theta) = f(\theta)$ if $\abs{f(\theta)} \leq n$ and $f_n(\theta)=0$ otherwise, where $(n, \theta) \in \bbN \times \Theta$.
Each $f_n$ is bounded, thus belongs to $L^q[\Theta, \mu]$, and the monotone convergence theorem applied to the functions $\abs{f_n}^q$ states that $\norm{f_n}_q \xrightarrow{n \to \infty} \norm{f}_q = +\infty$.
Thus, there exists $N \in \bbN$ such that $\norm{f_N}_q > 1$.
We can then apply \Cref{lemma:intermediate_fl} to get $g \in \mY$ and $C>0$ such that
\begin{equation*}
  \langle f_N, g \rangle_\mY - \norm{g}_p \geq C.
\end{equation*}
According to \Cref{lemma:intermediate_fl}, $g(\theta)=0$ whenever $f_N(\theta)=0$, which ensures that
\begin{equation*}
  \langle f, g \rangle_\mY = \langle f_N, g \rangle_\mY.
\end{equation*}
Taking a running parameter $t>0$, this means that
\begin{equation*}
  \langle f_N, t g \rangle_\mY -  \norm{tg}_p = \langle f, t g \rangle_\mY -  \norm{tg}_p \geq t C \xrightarrow{t \to \infty} +\infty,
\end{equation*}
which shows that $\norm{\cdot}^\star(f) = +\infty$.
\end{itemize}
\paragraph{Case $p=+\infty$:}
The reasoning goes as follows: we show that $\norm{f}_1 \leq 1$ implies $\norm{\cdot}_\infty^\star (f) = 0$, and that $\norm{f}_1 > 1$ gives $\norm{\cdot}_\infty^\star (f) = +\infty$, which allows one to conclude that $\norm{\cdot}_\infty ^\star = \iota_{\cB^1_1}(\cdot)$.
\begin{itemize}
\item \tb{When $\norm{f}_1 \leq 1$}: By applying Hölder's inequality we get that $ \langle f, g \rangle_\mY  \le
    \norm{f}_{1} \norm{g}_{\infty} \text{ for all } g \in \mY$. Using the condition that $\norm{f}_1 \leq 1$, this means that $\langle f, g \rangle_\mY -  \norm{g}_{\infty} \leq 0 \text{ for all } g \in \mY$.
Since the supremum is reached for $g=0$, we get that  $\norm{\cdot}_{\infty} ^\star (f) = 0$.
\item \tb{When $\norm{f}_1 > 1$}: Let $g \colon \theta \mapsto \sign{(f(\theta))}$.
Since $g$ is bounded by $1$, it belongs to $\mY$, and
$\langle f,g \rangle_\mY = \norm{f}_1$.
Running a free parameter $t>0$, this means that
$\langle f, t g \rangle_\mY - t \norm{g}_\infty = t \underbrace{(\norm{f}_1 - 1)}_{> 0} \xrightarrow{t \to \infty} +\infty$ which implies that $\norm{\cdot}_{\infty} ^\star (f) = +\infty$.
\end{itemize}
\qed
\subsection{Proof of \Cref{prop:huber_explicit}}
\label{sec:proof1}
\begin{proposition*}[\ref{prop:huber_explicit}]
  Let $\kappa > 0$, $p \in [1, +\infty]$, and $q$ the dual exponent of $p$ (i.e., $\frac{1}{p}+\frac{1}{q}=1$). Then for all $f \in \mY$,
  \begin{equation*}
  H_{\kappa}^p(f) =
  \left\{\begin{matrix*}
  \begin{array}{cl}
  \frac{1}{2} \|f\|_\mY^2 & \textrm{if } \norm{f}_q \le \kappa\\[0.15cm]
 \frac{1}{2} \norm{\Proj_{\cB_\kappa^q}(f)}_\mY^2 + \kappa \norm{f - \Proj_{\cB_\kappa^q}(f)}_p & \textrm{otherwise}.
  \end{array}
  \end{matrix*}\right..
  \end{equation*}
\end{proposition*}
\paragraph{Proof}
Let us introduce the notation $R(g) = \frac{1}{2} \norm{f-g}_\mY^2 + \kappa \norm{g}_p$ where $f \in \mY$, $g \in \mY$. Then
\begin{align}
    H_{\kappa}^p(f) & \stackrel{(a)}{=} \inf_{g \in \mY} R(g)
    \stackrel{(b)}{=} R(\prox_{\kappa \norm{\cdot}_p}(f))
    \stackrel{(c)}{=} \frac{1}{2} \norm{\Proj_{\cB_\kappa^q}(f)}_\mY^2 + \kappa \norm{f - \Proj_{\cB_\kappa^q}(f)}_p, \label{eq:Hkp}
\end{align}
where (a) follows from the definition of the infimal convolution,
(b) is implied by that of the proximal operator using that  $\kappa \norm{\cdot}_p \in \Gamma_0(\mY)$. (c) is a consequence of the Moreau decomposition (\Cref{lemma:moreau}) as
\begin{align}
   \prox_{\kappa \norm{\cdot}_p}(f) &= f - \prox_{\left(\kappa \norm{\cdot}_p\right)^\star}(f)
   \stackrel{(d)}{=} f - \prox_{\iota_{\cB_\kappa^q}}(f) \stackrel{(e)}{=} f -\Proj_{\cB_\kappa^q}(f),
   \label{eq:prox-p-norm->proj}
\end{align}
where in (d) and (e) we used that
\begin{align}
\left(\kappa \norm{\cdot}_p \right)^\star &\stackrel{(f)}{=} \iota_{\cB_\kappa^q}(\cdot)\text{ with $\frac{1}{p}+\frac{1}{q}=1$}, \label{eq:conjugate-of-scaled-norms} \\
\prox_{\iota_{\cB_\kappa^q}}&\stackrel{(g)}{=}\Proj_{\cB_\kappa^q}.  \label{eq:prox-of-scaled-balls}
\end{align}
(f) follows from the facts listed in the 3rd and the 2nd line of Table~\ref{tab:fl_useful}:
\begin{align*}
    \left(\kappa \norm{\cdot}_p \right)^\star &= \kappa \left( \norm{\cdot}_p \right)^\star(\cdot/\kappa) = \kappa \iota_{\cB_1^q}(\cdot/\kappa) = \iota_{\cB_\kappa^q}(\cdot).
\end{align*}
(g) is implied by $\iota_{\cB_\kappa^q} = \iota_{\cB_1^q}(\cdot/\kappa)$, the precomposition rule of proximal operators ($\prox_{f(\alpha \cdot )}=\frac{1}{\alpha} \prox_{\alpha^2f}(\alpha\cdot)$ holding for any $\alpha >0$; see (2.2) in \citep{parikh14proximal}),  and $\prox_{\iota_{\cB_1^q}}=\Proj_{\cB_1^q}$:
\begin{align*}
    \prox_{\iota_{\cB_\kappa^q}} &= \prox_{\iota_{\cB_1^q}(\cdot/\kappa)} =  \kappa \prox_{\frac{1}{\kappa^2} \iota_{\cB_1^q}}(\cdot/\kappa) = \kappa \prox_{\iota_{\cB_1^q}}(\cdot/\kappa)  = \kappa \Proj_{\cB_1^q}(\cdot/\kappa) = \Proj_{\cB_\kappa^q}.
\end{align*}
Finally we note that when $f= \Proj_{\cB_\kappa^q}(f)$ ($\Leftrightarrow f  \in \cB_\kappa^q \Leftrightarrow \norm{f}_q \le \kappa $), $\eqref{eq:Hkp}$ simplifies to $\frac{1}{2} \|f\|_\mY^2$.
\qed
\subsection{Proof of \Cref{prop:dual_huber}}
\label{sec:proof2}
\begin{proposition*}[\ref{prop:dual_huber}]
  Let $p, \kappa \in [1, +\infty] \times ]0, +\infty[$ and $\frac{1}{p}+\frac{1}{q}=1$.
  Then the dual of \Cref{pbm:primal_sup} with loss $H_\kappa^p$ writes as
  \begin{problem}
    \label{pbm:dual_sup_huber}
    \begin{aligned}
      \inf_{(\alpha_i)_{i\in [n]} \in \mY^n}
      &\sum_{i\in [n]} \frac{1}{2}\norm{\alpha_i}_{\mY}^2 - \langle \alpha_i, y_i \rangle_\mY
       + \frac{1}{2 \lambda n} \sum_{i,j\in [n]}^n  k_{\mX}(x_i, x_j) \left \langle \alpha_i, T_{k_\Theta} \alpha_j \right \rangle_\mY  \text{ s.t. } \norm{\alpha_i}_q \leq \kappa \text{ for  } \forall i \in [n].
    \end{aligned}
  \end{problem}
\end{proposition*}
\paragraph{Proof}
Applying \Cref{lemma:dual_pbm_sup} and \eqref{eq:conjugate-of-scaled-norms} give the result.
\qed
\subsection{Proof of \Cref{prop:projection_q_ball}}
\label{sec:proof3}
\begin{proposition*}[\ref{prop:projection_q_ball}]
  Let $\kappa > 0$.
  The projection on $\cB_{\kappa}^q$ is tractable for $q=2$ and $q=\infty$ and can be expressed for all $(\alpha, \theta) \in \mY \times \Theta$ as
  \begin{align}
    \label{eq:proj_norm_2_sup}
    \Proj_{\cB_\kappa^2}(\alpha) &= \min{\left(1, \frac{\kappa}{\norm{\alpha}_2}\right)} \alpha, \quad \text{if } \alpha \neq 0\\
    \label{eq:proj_norm_inf_sup}
    \left(\Proj_{ \cB_\kappa^\infty }(\alpha) \right)(\theta) &= \sign{(\alpha(\theta))} \min{(\kappa, |\alpha(\theta)|)}.
  \end{align}
\end{proposition*}
\paragraph{Proof}
The projection on the $2$-ball of radius $\kappa$ is similar to the finite-dimensional case $(\mY = \R^d)$ for which \Cref{eq:proj_norm_2_sup} is well-known.

We now turn to the case of $q=\infty$. Let $\alpha \in \mY$. By definition,
\begin{equation}
  \label{eq:proj_inf}
  \Proj_{ \cB_\kappa^\infty }(\alpha) = \argmin_{y \in \mY} \frac{1}{2} \int_{\Theta} \left[ \alpha(\theta) - y(\theta) \right]^2 \mathrm{d}\mu(\theta) + \iota_{\cB_\kappa^\infty}(y).
\end{equation}
Since $\alpha \in \mY$, the function $g$ defined as
\begin{equation*}
  g(\theta) = \sign{(\alpha(\theta))} \min{(\kappa, |\alpha(\theta)|)},\quad \theta \in \Theta,
\end{equation*}
is measurable, it is in $\mY$, and one can verify  easily that it is the solution of \Cref{eq:proj_inf}; it corresponds to taking the pointwise projection of $\alpha$ on the segment $[-\kappa, \kappa]$.
\qed
\subsection{Proof of \Cref{prop:l_eps_p}}
\label{sec:proof4}
\begin{proposition*}[\ref{prop:l_eps_p}]
  Let $\epsilon > 0$ and $p \in [1, +\infty]$. Then
  $
  \ell_\epsilon^p(f) = \frac{1}{2}\norm{f -\Proj_{\cB_\epsilon^p}(f)}^2_\mY
  $ for all $f \in \mY$.
\end{proposition*}
\paragraph{Proof}
Let $R(g) = \frac{1}{2} \norm{f-g}_\mY^2 + \iota_{\cB_\epsilon^p}(g)$ where $f \in \mY$ and  $g \in \mY$.
Then
\begin{align*}
    \ell_\epsilon^p(f) & \stackrel{(a)}{=} \inf_{g \in \mY} R(g) \stackrel{(b)}{=}R\left(\prox_{\iota_{\cB_\epsilon^p}}(f)\right) \stackrel{(c)}{=} R\left(\Proj_{\cB_\epsilon^p}(f)\right) \stackrel{(d)}{=} \frac{1}{2}\norm{f -\Proj_{\cB_\epsilon^p}(f)}^2_\mY,
\end{align*}
where (a) follows from the definition of the infimal convolution, (b) is implied by that of the proximal operator and by $\iota_{\cB_\epsilon^p} \in \Gamma_0(\mY)$, (c) is the consequence of $\prox_{\iota_{\cB_\epsilon^p}}(f) = \Proj_{\cB_\epsilon^p}(f)$ implied by \eqref{eq:prox-of-scaled-balls}, in (d) the definition of $R$ was applied.
\qed
\subsection{Proof of \Cref{prop:dual_eps}}
\label{sec:proof5}
\begin{proposition*}[\ref{prop:dual_eps}]
  Let $(p, \epsilon) \in [1, +\infty] \times ]0, +\infty[$, and $\frac{1}{p}+\frac{1}{q}=1$. Then, the dual of \Cref{pbm:primal_sup} writes as
  \begin{problem}
    \label{pbm:dual_sup_epsilon}
    \begin{aligned}
      \inf_{(\alpha_i)_{i\in [n]} \in \mY^n} &
      \sum_{i\in [n]} \left[\frac{1}{2}\norm{\alpha_i}_{\mY}^2 - \langle \alpha_i, y_i \rangle_\mY + \epsilon \norm{\alpha_i}_q\right]
      + \frac{1}{2 \lambda n} \sum_{i,j\in [n]} k_{\mX}(x_i, x_j) \left \langle \alpha_i, T_{k_\Theta} \alpha_j \right \rangle_\mY.
    \end{aligned}
  \end{problem}
\end{proposition*}
\paragraph{Proof}
Applying \Cref{lemma:dual_pbm_sup} with
\begin{align}
    \left(\iota_{\cB_\epsilon^p}(\cdot)\right)^\star = \epsilon \norm{\cdot}_q \label{eq:fenchel-indicator}
\end{align}
gives the result. \eqref{eq:fenchel-indicator} is the consequence of \eqref{eq:conjugate-of-scaled-norms} and the involutive property of the Fenchel-Legendre conjugate.
\qed

\subsection{Proof of \Cref{prop:proximal_q_norm}}
\label{sec:proof6}
\begin{proposition*}[\ref{prop:proximal_q_norm}]
Let $\epsilon > 0$. The proximal operator of $\epsilon \norm{\cdot}_q$ is computable for $q=1$ and $q=2$, and given for all $(\alpha, \theta) \in \mY \times \Theta$ by
\begin{align}
  \label{eq:prox_norm_1_sup}
  \left(\prox_{ \epsilon \norm{\cdot}_1 }(\alpha) \right)(\theta) &= \sign{(\alpha(\theta))} \left| \left|\alpha(\theta)\right| - \epsilon \right|_+, \\
  \label{eq:prox_norm_2_sup}
  \prox_{\epsilon \norm{\cdot}_2 }(\alpha) &= \alpha
  \left | 1 - \frac{ \epsilon}{\norm{\alpha}_\mY} \right |_+ \quad \text{if } \alpha \neq 0.
\end{align}
\end{proposition*}
\paragraph{Proof}
By \eqref{eq:prox-p-norm->proj} we know that
\begin{equation}
  \label{eq:prox_proj}
  \prox_{\epsilon \norm{\cdot}_q}(f) = f - \Proj_{\cB_\epsilon^p}(f).
\end{equation}
The projection operator is known from \Cref{prop:projection_q_ball} in the case of $p=2$ and $p=\infty$, which allows to express the proximal operator of the $q$-norm for $q=2$ and $q=1$ and by substituting \Cref{eq:proj_norm_2_sup} and \Cref{eq:proj_norm_inf_sup} into \Cref{eq:prox_proj}.
\qed

\section{Additional Details}
\label{sec:expes_supplement}
In this section we present additional algorithmic details as well as complement the numerical experiments presented in the main document.
\subsection{Algorithmic Details}
\Cref{alg:pgd_eigen} fully describes how to learn models with the representation relying on the eigendecomposition of the integral operator developed in \Cref{sec:huber_eigen} and \Cref{sec:epsilon_eigen}.

\LinesNumberedHidden{
\begin{algorithm}[t]
\SetKwInOut{Input}{input}
\SetKwInOut{Init}{init}
\SetKwInOut{Parameter}{Param}
\SetInd{0.5em}{0.5em}
\caption{APGD with eigenbasis representation
\label{alg:pgd_eigen}}
\Input{~Gram matrix $\b{K}_\mX$, matrix of eigenvalues $\b{\Delta}$, data scalar product matrix $\b{R}$, regularization parameter $\lambda$, loss parameters $(\kappa, 2)$ or $(\epsilon, 2)$, gradient step $\gamma$}\vspace{0.1cm}
\Init{ $\b{A}^{(0)}, \b{A}^{(-1)} = \bm{0} \in \mathbb{R}^{n \times r}$}
\For{epoch $t$ from $0$ to $T-1$}{
    \tcp{gradient step}
    $\b{V} = \b{A}^{(t)} + \frac{t - 2}{t + 1} \left (\b{A}^{(t)} - \b{A}^{(t-1)}  \right)$

    $\b{U} = \b{V} - \gamma\left( \b{V} + \frac{1}{\lambda n}\b{K}_\mX \b{V} \b{\Delta} - \b{R} \right)$

    \tcp{proximal step}
    \For{row $i \in [n]$}{
        $\b{a}_{i}^{(t+1)} = \min\left(\frac{\kappa}{\norm{\b{u}_{i}}_2}, 1\right) \b{u}_{i}$
        \tcp{if $H_\kappa^2$}
        $\b{a}_{i}^{(t+1)} =
        \left | 1 - \frac{\gamma \epsilon}{\norm{\b{u}_{i}}_2} \right |_+ \b{u}_i$
        \tcp{if $\ell_\epsilon^2$}
        }
    }
\Return{$\b{A}^{(T)}$}
\end{algorithm}}

\subsection{Synthetic Data}
\label{sec:synth}
Below we detail the generation process of the synthetic dataset (Section~\ref{sec:gen-process}), we expose in full detail the parameters used in the experiments (Section~\ref{sec:exp-details}; see Fig.~\ref{fig:toyeps} and \ref{fig:toyout} in the main paper), and we provide additional illustration for the interaction between the Huber loss' $\kappa$ and the regularization parameter $\lambda$ (Section~\ref{sec:extra-illustrations}).

\subsubsection{Generation Process} \label{sec:gen-process}

Given covariance parameters $(\pmb \sigma^{\text{in}}, \pmb \sigma^{\text{out}}) \in \mathbb R^r \times \mathbb R^r$ for $c \in [r]$ we draw and fix Gaussian processes $g_c^{\text{in}} \sim \mathcal{GP}(0, k_{\sigma^{\text{in}}_c})$ and $g_c^{\text{out}} \sim \mathcal{GP}(0, k_{\sigma^{\text{out}}_c})$.
\begin{figure}
  \begin{center}
    \includegraphics[width=0.50\textwidth]{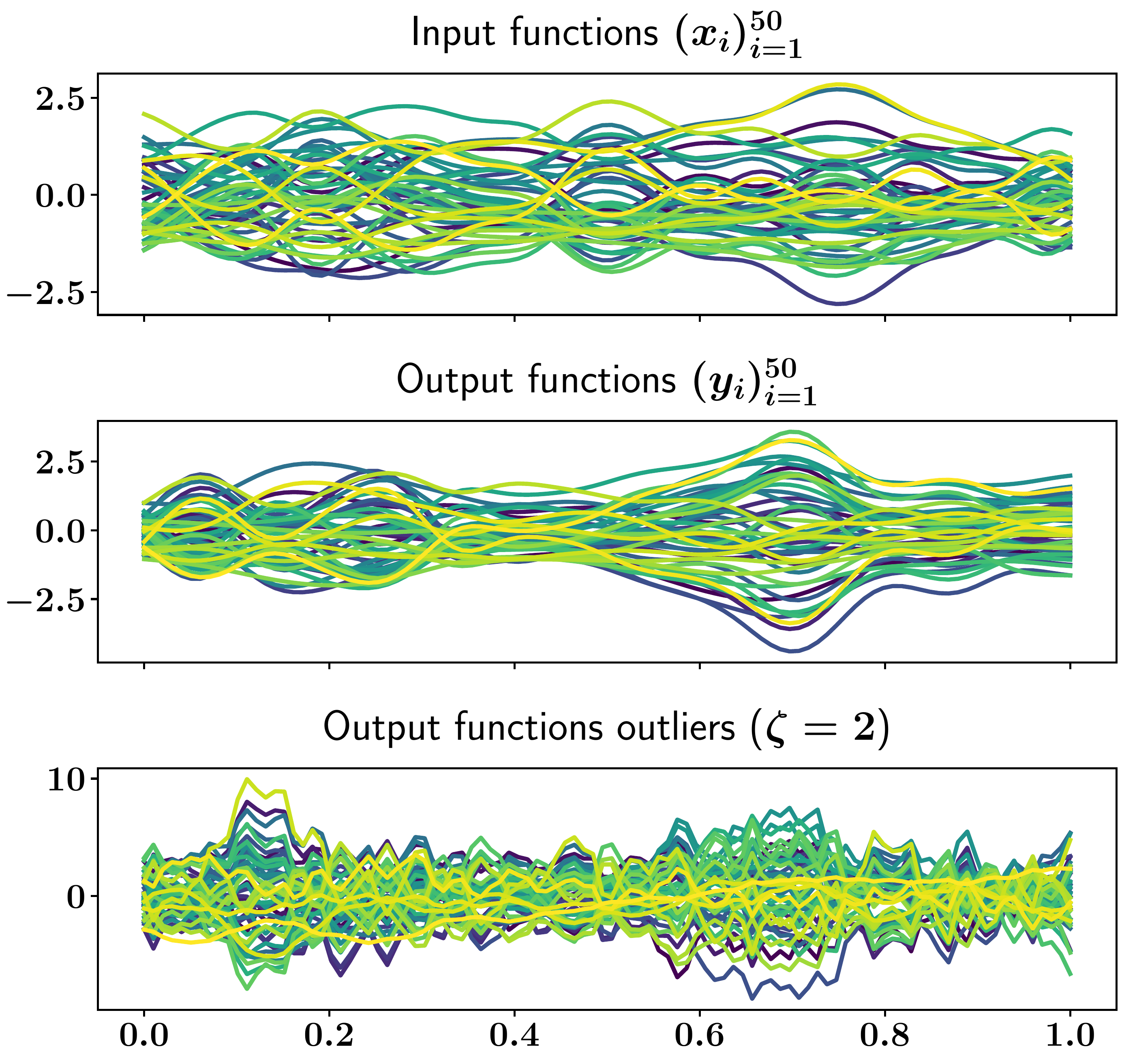}
  \end{center}
  \caption{Examples from the toy dataset and corresponding type 2 outliers.}
  \label{fig:toyexs}
\end{figure}
We then generate $n$ samples as $\left(\sum_{c\in [r]} u_{ic} g_c^{\text{in}}, \sum_{c\in [r]} u_{ic} g_c^{\text{out}}
\right)_{i\in [n]}$, where the coefficients $u_{ic}$ are drawn i.i.d. according to a uniform distribution $\mathcal U([-0.5, 0.5])$. In the experiments, we take $r=4$ and set $\pmb \sigma^{\text{in}} = \pmb \sigma^{\text{out}} = (0.05, 0.1, 0.5, 0.7)$. We show input and output functions drawn in this manner in the first and second row of Fig.~\ref{fig:toyexs}. In the bottom row we display outliers of Type 2 with $\pmb \sigma = (0.01, 0.05, 1, 4)$ and intensity $\zeta=2$. For the contaminated indices $i$ in $I$ we add the corresponding outlier to the function $y_i$

\subsubsection{Experimental Details} \label{sec:exp-details}
We provide here the full details of the parameters used for the experiments on the toy dataset. For all experiments, we fix the parameter $\rho^{\text{in}}$ of the input Gaussian kernel $k_\mX: (x_1, x_2) \longmapsto \exp\left(- \rho \|x_0 - x_1 \|_{\mX}^2\right)$ to $\rho^{\text{in}}=0.01$ and that of the output Gaussian kernel to  $\rho^{\text{out}}=100$. Indeed, since we are only given discrete observations for the input functions as well, we use the available observations to approximate the norms in the above kernels. For the experiments on robustness which results are displayed in Fig.~\ref{fig:toyout} of the main paper, we select via cross-validation the regularization parameter $\lambda$ and the $\kappa$ parameters of the Huber loss, considering values in a geometric grid of size $10$ ranging from $10^{-6}$  to $10^{-3}$ for $\lambda$ and values in a geometric grid of size $25$ ranging from $10^{-3}$ to $10^{-1}$ for $\kappa$.

\subsubsection{Additional Illustrations}\label{sec:extra-illustrations}
\begin{figure}
     \begin{center}
     \begin{subfigure}[b]{0.7\textwidth}
        \centering
		\includegraphics[width=\linewidth]{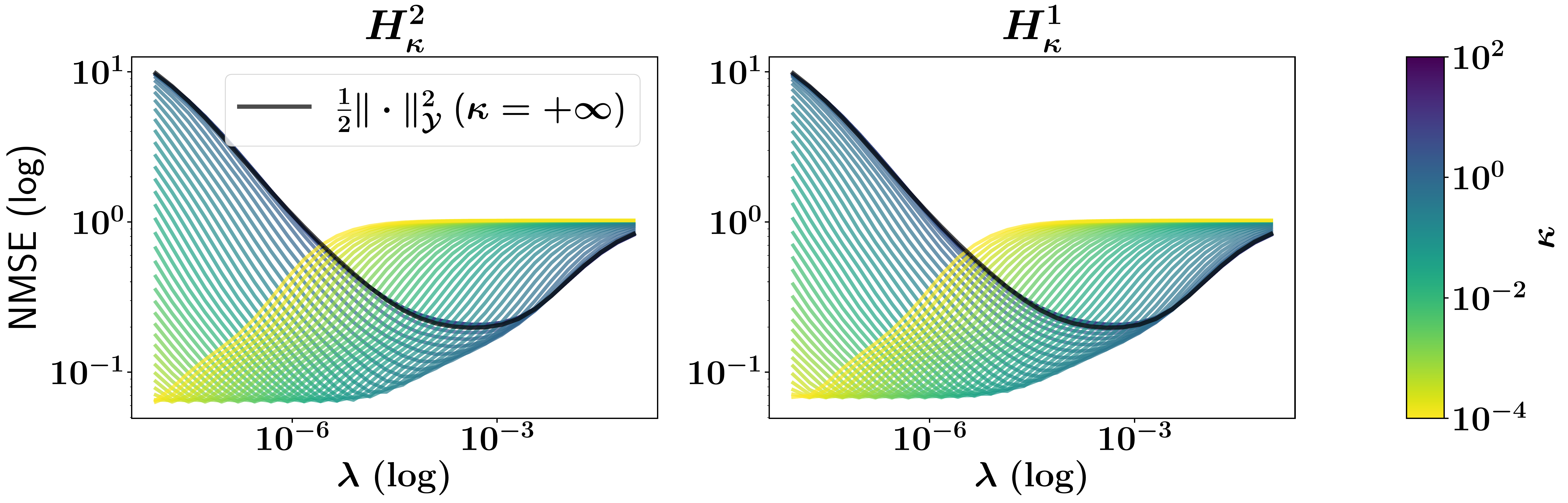}
		\caption{Type 2 outliers with $\tau=0.2$ and $\zeta=2$}
		\label{fig:hubdescglob}
     \end{subfigure}
     \begin{subfigure}[b]{0.7\textwidth}
        \centering
        \includegraphics[width=\linewidth]{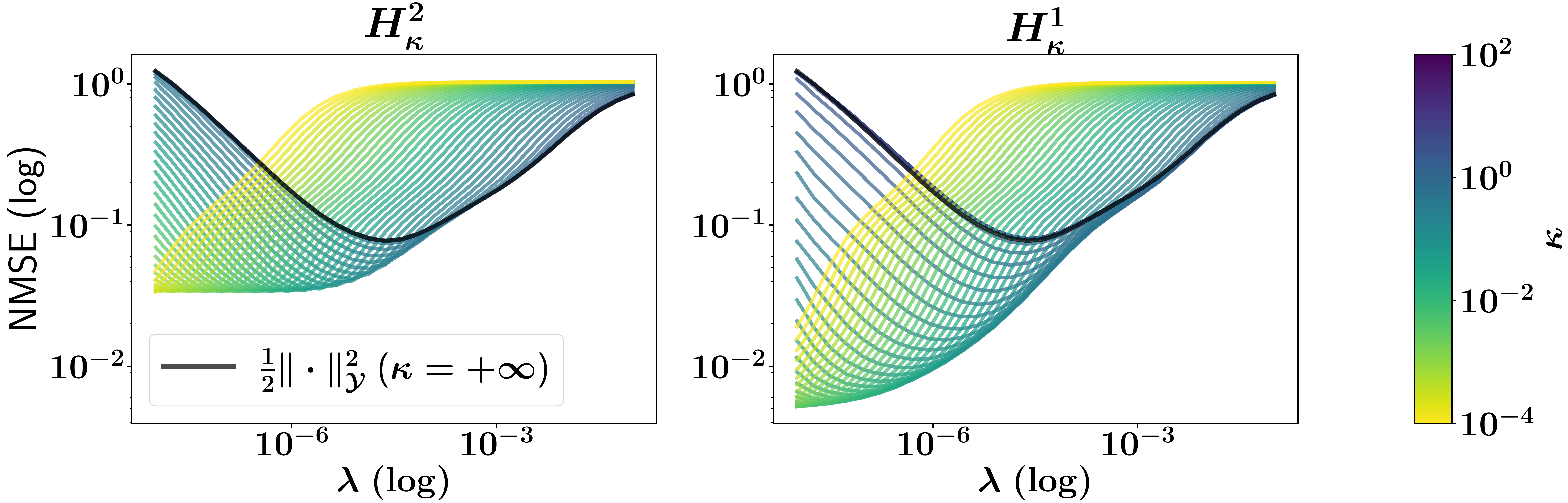}
	    \caption{Type 3 outliers with $\tau=0.3$ and $\xi=0.3$}
	    \label{fig:hubdescloc}
     \end{subfigure}
    \caption{NMSE as a function of $\lambda$ and Huber losses' $\kappa$ with two types of outliers.}
    \label{fig:hubdesc}
    \end{center}
\end{figure}

To highlight the interaction between the regularization parameter $\lambda$ and the parameter $\kappa$ of the Huber loss, we plot the NMSE values for various values of $\lambda$ and $\kappa$ using the toy dataset corrupted with the two main types of contamination used in the main paper, Type 2 and Type 3 outliers. The results are displayed in Fig.~\ref{fig:hubdesc} and confirm that by making $\kappa$ and $\lambda$ vary, when the data are corrupted, we can always find a configuration that is significantly more robust than the square loss. In accordance with one's expectation, when dealing with local outliers (Fig.~\ref{fig:hubdescloc}), the loss $H_\kappa^1$ is much more efficient than the loss $H_\kappa^2$. However, when dealing with global outliers (Fig.~\ref{fig:hubdescglob}), the two losses perform equally well.

\subsection{DTI Data}
In this section we provide details regarding the experiments on the DTI dataset. For this dataset, we use a Gaussian kernel as input kernel and a Laplace kernel as output kernel, for the first we fix its parameter to $\rho^{\text{in}} = 1.25$, and for the second, defined as $k_{\Theta}: (\theta_1, \theta_2) \longmapsto \exp (-\rho^{\text{out}} \|x_0 - x_1 \|_{\mX} )$, we fix its parameter to $\rho^{\text{out}}=10$. We consider two values of $\lambda$, the first one ($\lambda = 10^{-5}$) is chosen too small for the square loss to highlight the additional sparsity-inducing regularization possibilities offered by the $\epsilon$-insensitive loss through the parameter $\epsilon$, while the second one ($\lambda=10^{-3}$) corresponds to a near-optimal value for the square loss. We do cross-validate the parameters of the losses. For the loss $\ell_\epsilon^2$ we consider values of $\epsilon$ in a geometric grid of size 50 ranging from $10^{-3}$ to $10^{-1}$, while for the loss $\ell_\epsilon^\infty$, we search in a geometric grid of the same size, however ranging from $10^{-3}$ to $10^{-0.5}$. For the Huber losses $H_\kappa^1$ and $H_\kappa^2$, we search for $\kappa$ using a geometric grid of size 50 ranging this time from $10^{-4}$ to $10^{-1}$.

\subsection{Speech Data}
\label{sec:speech_sup}
This section is dedicated to additional details about the experiments carried out on the speech benchmark.

\noindent {\bf Input kernel:} As highlighted in the main paper, we encode the input sounds through 13 mel-frequency cepstral coefficients (MFCC). To deal with this particular input data type we used the following kernel. Let $((x_{ij}^{(v)})_{v\in [13]})_{i\in [n],j\in[m]}$ be the transformed input data where $v$ serves as an index for the MFCC number. The number of locations $m$ is the same for all $i \in [n]$ since we extend the signals to match the longest one to be able to train the models. We then center and reduce each MFCC using all samples and sampling locations to compute the mean and standard deviation; let $((\tilde x_{ij}^{(v)})_{v\in[13]})_{i\in[n], j\in [m]}$ be the resulting standardized input data. Denoting by $\tilde x_i$ the element $((\tilde x_{ij}^{(v)})_{v\in[13]})_{j\in[m]}$, we then use the following kernel:
$$ k_\mX: (\tilde x_0, \tilde x_1) \longmapsto \frac 1 m \sum_{j\in [m]} \exp \left ( -\rho^{\text{in}} \sum_{v\in [13]} \left (\tilde x_{0j}^{(v)} - \tilde x_{1j}^{(v)} \right )^2 \right ).$$

\noindent {\bf Experimental details:} For all the experiments (with or without corruption), we select the parameter of the input kernel $\rho^{\text{in}}$, the regularization parameter and the parameters of the losses  using cross-validation. We fix the parameter of the Laplace output kernel to $\rho^{\text{out}}=10$. However, to reduce the computational burden, we perform the selection of the parameter $\rho^{\text{in}}$ only for the square loss, and then take the corresponding values for the other losses. For this parameter values in a geometric grid of size $15$ ranging from $10^{-2}$ to $10^{-0.5}$ are considered. For $\lambda$, the search space is a  geometric grid of size $10$ ranging from $10^{-10}$ to $10^{-6}$. Finally, for the $\epsilon$-insensitive loss, values of $\epsilon$ in a geometric grid of size $80$ ranging from $10^{-5}$ to $10^{-1}$ are considered, while for the Huber losses we search for $\kappa$ in a geometric grid of size $100$ ranging from $10^{-7}$ to $1$.

\section{Illustration of Loss Functions}
\label{sec:loss_plots}
\label{sec:loss_illustration}
\begin{figure}
  \begin{center}
    \includegraphics[width=0.7\linewidth]{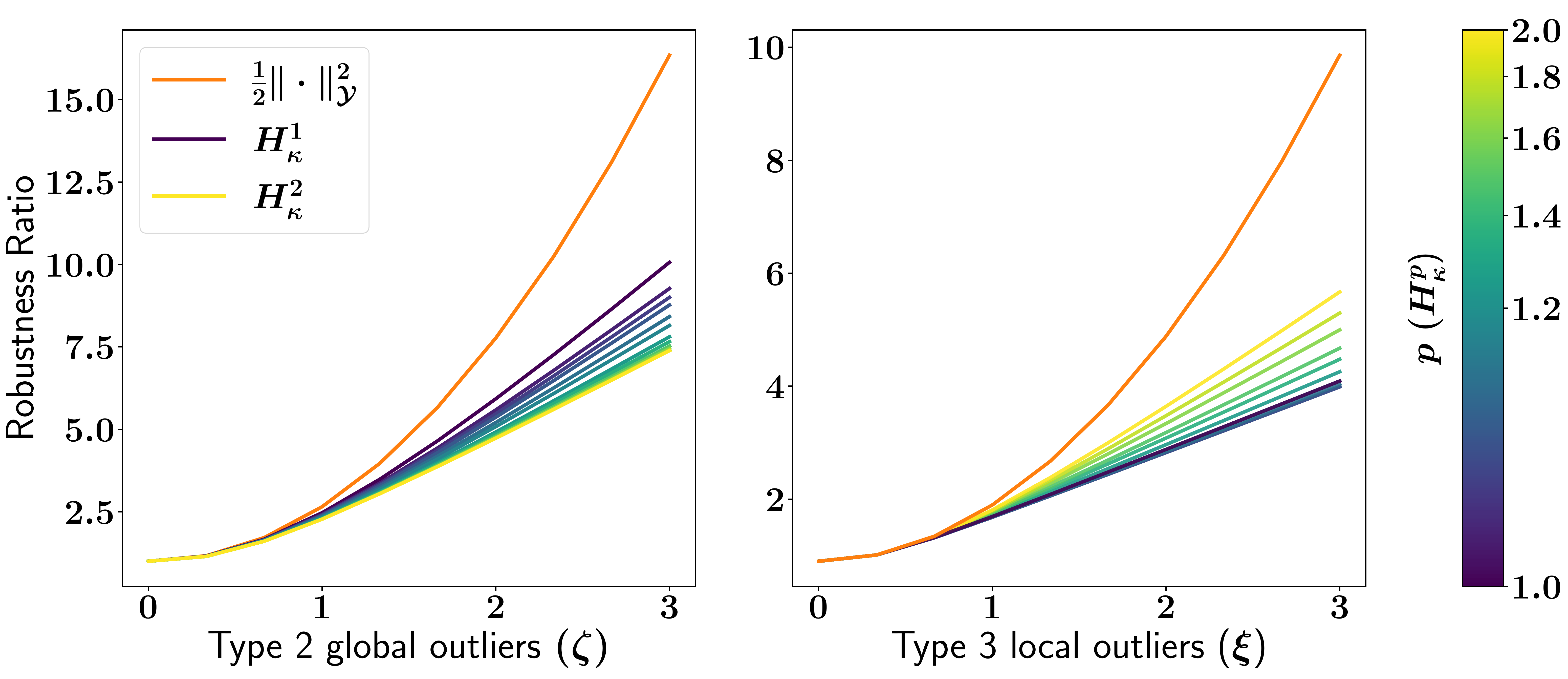}
  \end{center}
  \caption{Sensitivity of $H_\kappa^p$ to outliers for various $p \in [1, 2]$}
  \label{fig:pscomp}
\end{figure}

In this section we illustrate the differences between our proposed convoluted losses in several ways. In Section \ref{subsec:discussp} we study empirically how the choice of $p$ affects the sensitivity of the Huber loss $H_\kappa^p$ to different kind of outliers. In Section \ref{subsec:lossex}, we plot some of our proposed losses when they are defined either on $\mathbb R$ or $\mathbb R^2$.

\subsection{Discussion on the Choice of $p$ for $H_\kappa^p$} \label{subsec:discussp}

As it is highlighted in the main paper, solving Problem (\ref{pbm:dual_huber_splines}) for $p \not \in \{1, 2 \}$ is unpractical since it involves the computation of $n$ projections on a $q$-ball at each APGD iteration. Performing  such projection is feasible (it is a convex optimization problem) but it has to be done in an iterative way. In our case, to run APGD with such inner iterations turns out to be too time consuming. However we still can approximately calculate the losses $H_\kappa^p$ using Proposition \ref{prop:huber_explicit} for any $p$ (computing the involved projection iteratively). We thus propose to leverage this possibility to study empirically the sensitivity of the Huber losses $H_\kappa^p$ to global and local outliers, for different values of $p$.

The impact of the outliers on the solution of a regularized empirical risk minimization problem is partly determined by the contribution of the outliers to the data-fitting term relatively to the contribution of the normal observations. In order to investigate this aspect, we study and define next a quantity which we call Robustness Ratio.

Let $(e_i)_{i \in [n]} \in (L^2[\Theta, \mu])^n$ be a set of functional residuals and let $(\tilde e_i)_{i\in [n]}$ be the same functional residuals but contaminated with outliers. In practice, we have to choose a probability distribution to draw the functions $(e_i)_{i\in [n]}$ from, and an outlier distribution to corrupt those. For the functions $(e_i)_{i\in [n]}$ we use our synthetic data generation process (see Section \ref{sec:gen-process}), and for the outliers, we consider the same type 2 and type 3 outliers as in the experiments in Section \ref{subsec:toy} from the main paper. We then define the Robustness Ratio as 
\begin{align*}
    \text{Robustness Ratio} &:= \inf_{\kappa \geq 0} \frac 1 n \sum_{i\in [n]} \frac{H_\kappa^p (\tilde e_i)}{H_\kappa^p (e_i)}.
\end{align*}
The best value of this quantity is $1$; it means that the loss is not affected at all by the outliers, but it is indeed not possible to reach such value. In practice, we restrain our study to $p \in [1, 2]$. For each $p$ we reduce the search for $\kappa$ to different empirical quantiles of the $q$-norms of the uncorrupted functions $(e_i)_{i\in [n]}$, where $q$ is the dual exponent of $p$. It makes sense to do so since $\kappa$ corresponds to a $q$-norm threshold which separates observations considered to be outliers from those deemed normal (see Proposition \ref{prop:huber_explicit}). We consider the $\{0.5, 0.6, 0.7, 0.8, 0.9, 0.95, 0.99 \}$-th such empirical quantiles. For each $p$, we compute the robustness ratio for $\kappa$ equal to each of those quantiles, and then for each level of corruption, we select the value which minimizes the ratio. This indeed corresponds to an ideal setting, since in practice, we never have access to the uncorrupted data and we can never optimize $\kappa$ in this way. Thus the robustness ratio reflects more of a general robustness property of the loss in an optimal setting.

In accordance with one's expectation, when the data is contaminated with global outliers (left panel of Fig.~\ref{fig:pscomp}), it is better to choose $p=2$ whereas when the contamination is local (right panel of Fig.~\ref{fig:pscomp}), $p=1$ is almost the best choice; even though it seems that choosing $p$ slightly bigger than $1$ could be a tad better. Even though, we highlight that this analysis based on the Robustness Ratio has its limits; indeed we do not take into account the interplay between the data-fitting term and the regularization term which takes place during optimization. This certainly explains why we found the losses $H_\kappa^1$ and $H_\kappa^2$ to perform equally well in practice whereas based only on the Robustness Ratio analysis (left panel of Fig.~ \ref{fig:pscomp}) we would have said otherwise. The findings in presence of local outliers (right panel of Fig~. \ref{fig:pscomp}) are nevertheless coherent with what we observed in practice for the losses $H_\kappa^1$ and $H_\kappa^2$ in our experiments.

\subsection{Loss Examples in 1d and 2d} \label{subsec:lossex}

\begin{figure}
     \begin{center}
     \begin{subfigure}[b]{0.35\textwidth}
        \centering
		\includegraphics[width=\linewidth]{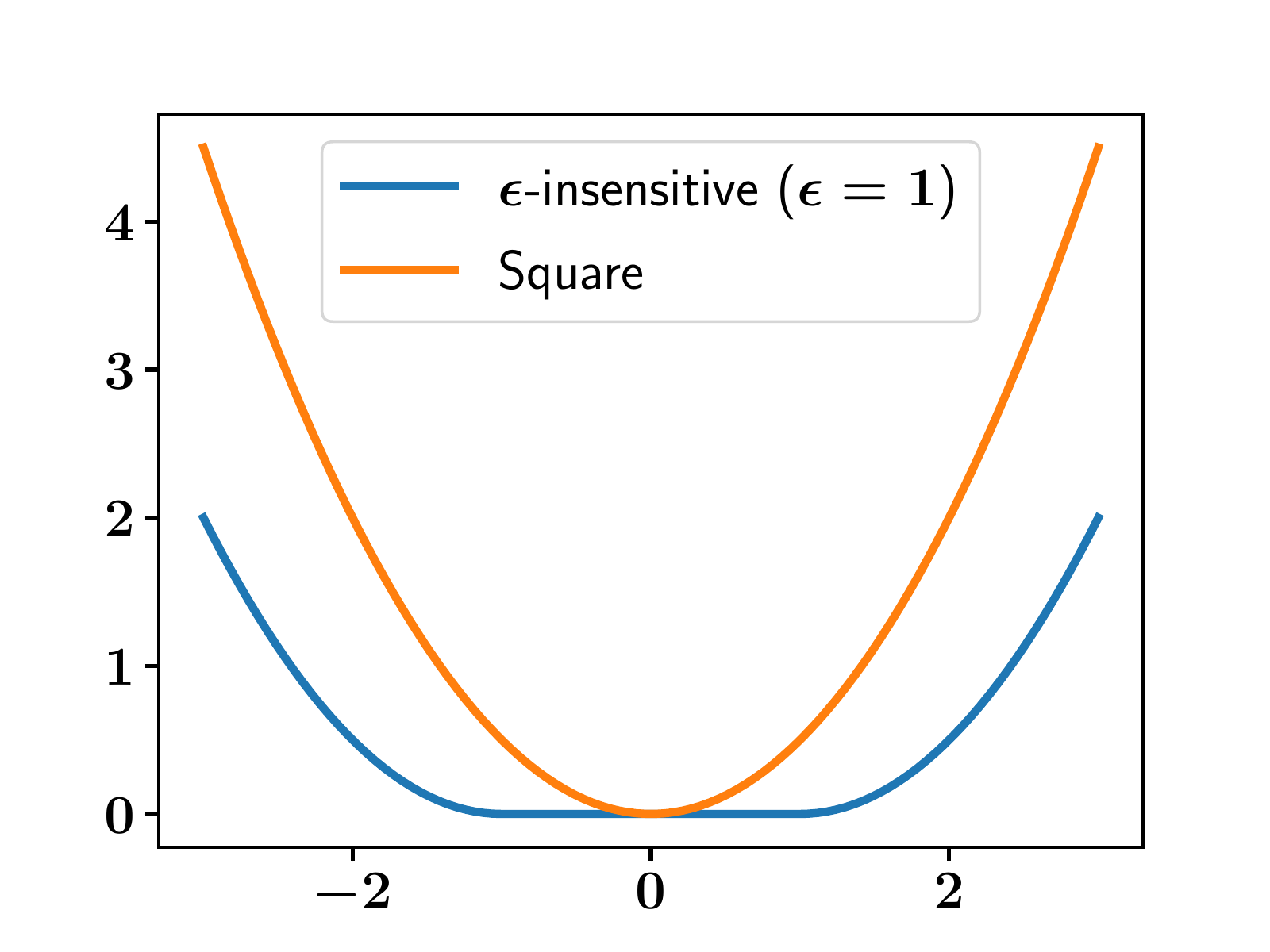}
		\caption{$\epsilon$-insensitive ($\epsilon=1$) and square loss}
		\label{fig:eps1d}
     \end{subfigure}
     \hspace{1.25cm}
     \begin{subfigure}[b]{0.35\textwidth}
        \centering
        \includegraphics[width=\linewidth]{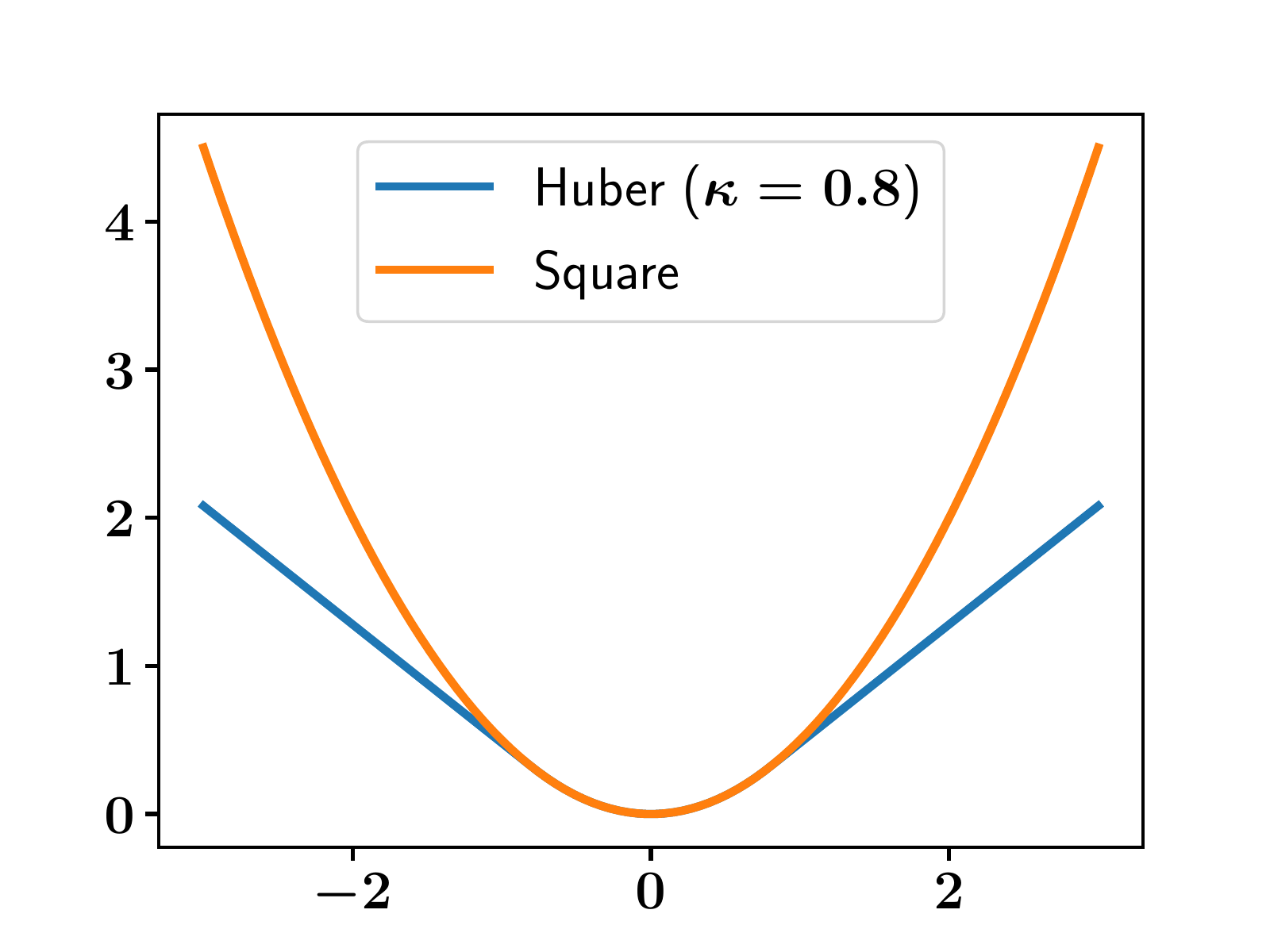}
        \caption{Huber loss ($\kappa=0.8$) and square loss}
	    \label{fig:hub1d}
     \end{subfigure}
     \end{center}
     \caption{Illustrations of the different losses defined on $\mathbb R$.}
    \label{fig:losses1d}
\end{figure}

In this section, we plot several of the proposed losses when they are defined on $\mathbb R$ and $\mathbb R^2$. In Fig.~\ref{fig:losses1d}, we compare the Huber (Fig.~\ref{fig:hub1d}) and the $\epsilon$-insensitive (Fig.~\ref{fig:eps1d}) losses with the square loss when they are defined on $\mathbb R$. 

\begin{figure}
     \begin{center}
     \begin{subfigure}[b]{0.44\textwidth}
         \centering
        \includegraphics[width=\linewidth]{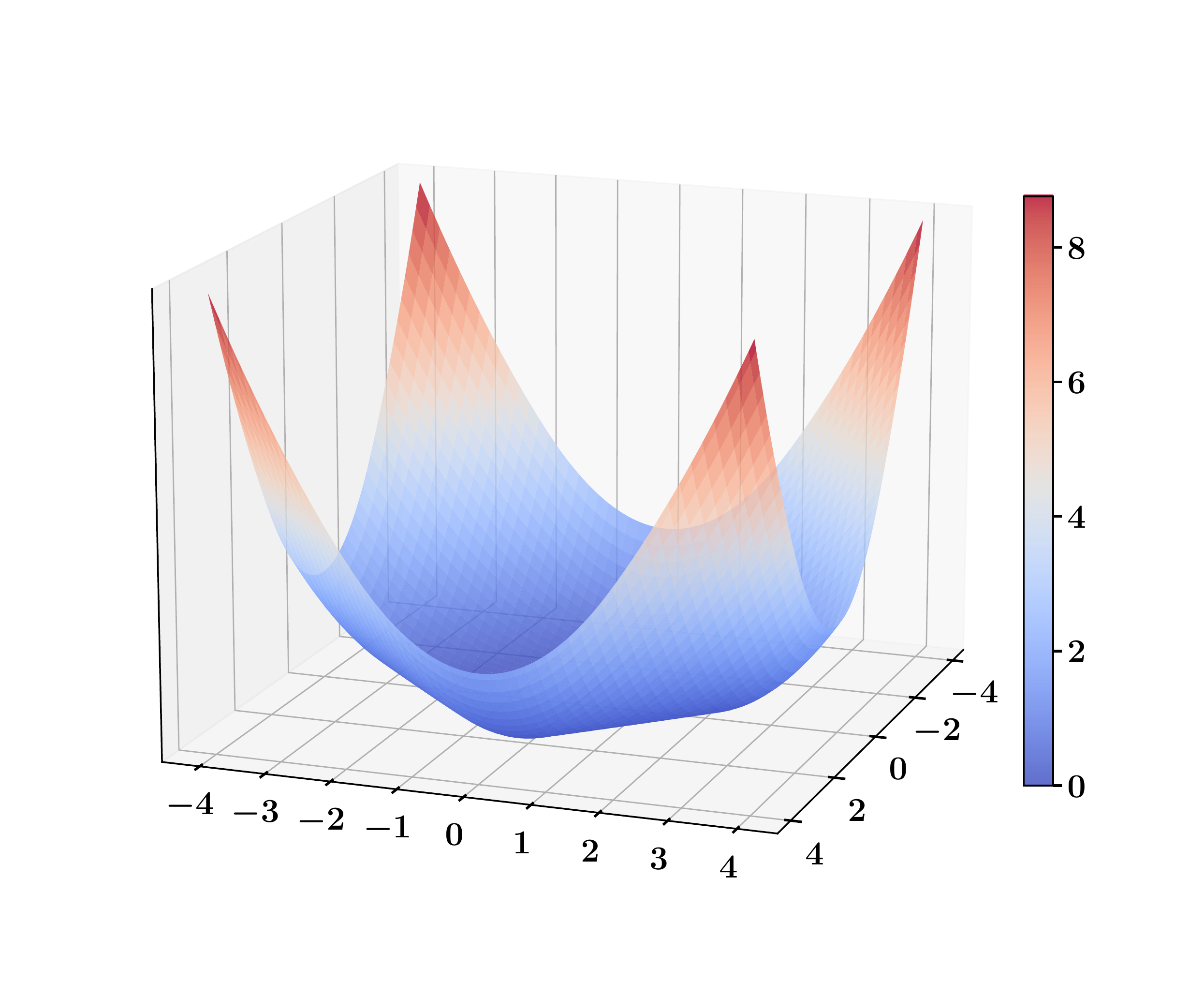}
        \caption{$\ell_{\epsilon}^{1.01}$ ($\epsilon=1$)}
        \label{fig:eps1012d}
     \end{subfigure}
     \begin{subfigure}[b]{0.44\textwidth}
         \centering
        \includegraphics[width=\linewidth]{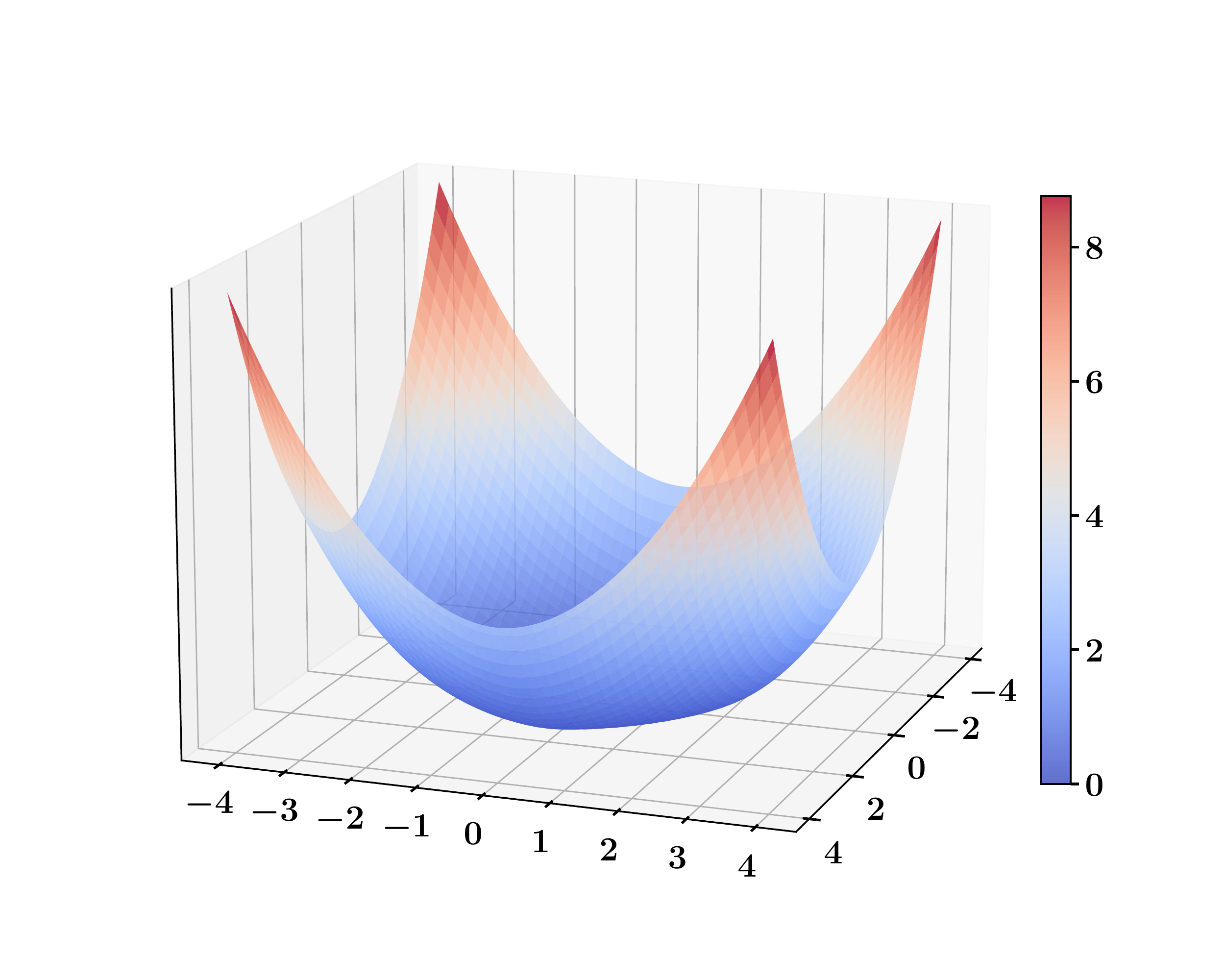}
        \caption{$\ell_{\epsilon}^{1.5}$ ($\epsilon=1$)}
        \label{fig:eps152d}
     \end{subfigure}
     \begin{subfigure}[b]{0.47\textwidth}
         \centering
        \includegraphics[width=\linewidth]{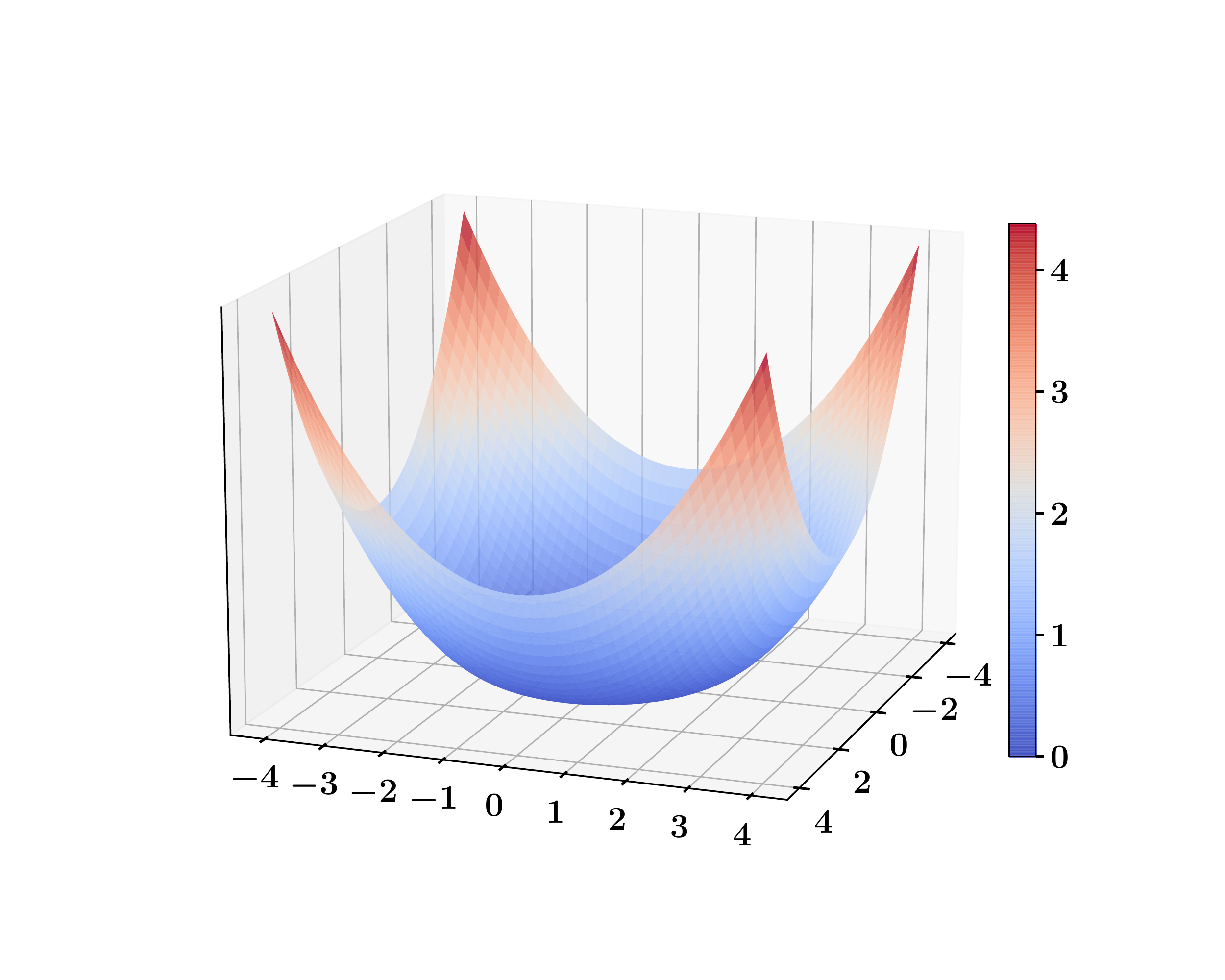}
        \caption{$\ell_{\epsilon}^2$ ($\epsilon=1$)}
        \label{fig:eps22d}
     \end{subfigure}
     \begin{subfigure}[b]{0.44\textwidth}
         \centering
        \includegraphics[width=\linewidth]{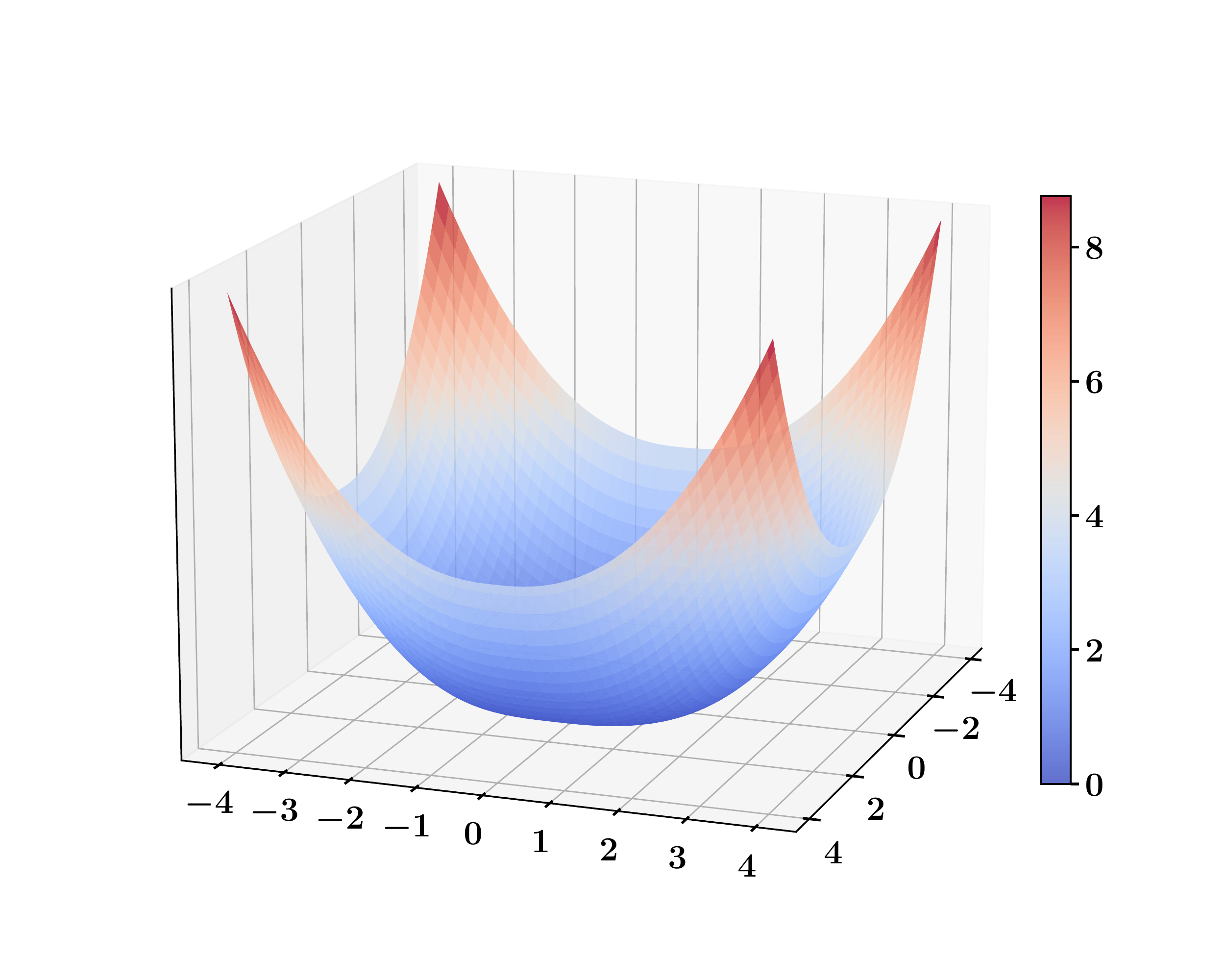}
        \caption{$\ell_{\epsilon}^3$ ($\epsilon=1$)}
        \label{fig:eps32d}
     \end{subfigure}
     \begin{subfigure}[b]{0.44\textwidth}
         \centering
        \includegraphics[width=\linewidth]{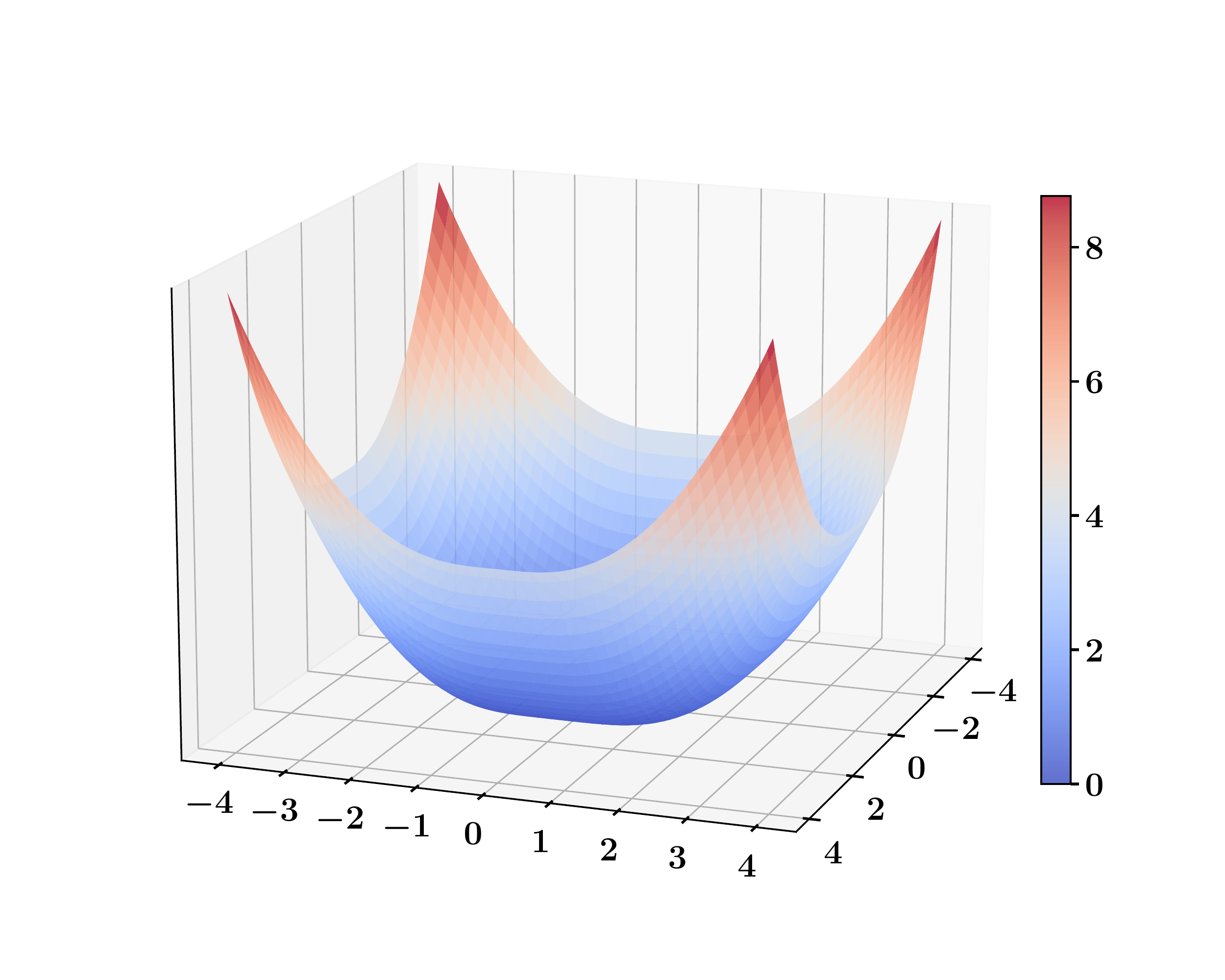}
        \caption{$\ell_{\epsilon}^5$ ($\epsilon=1$)}
        \label{fig:eps52d}
     \end{subfigure}
     \begin{subfigure}[b]{0.47\textwidth}
         \centering
        \includegraphics[width=\linewidth]{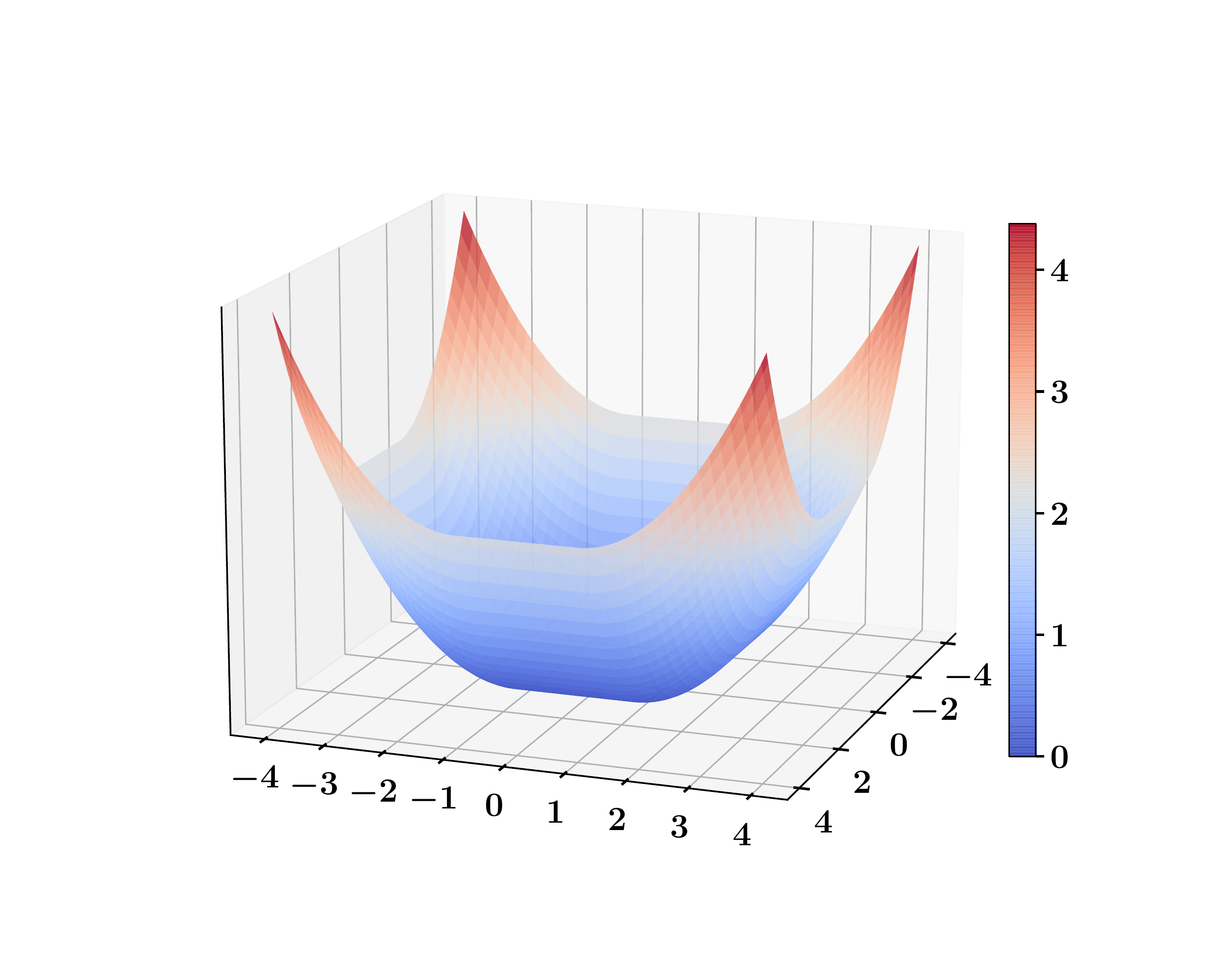}
        \caption{$\ell_{\epsilon}^\infty$ ($\epsilon=1$)}
        \label{fig:epsinf2d}
     \end{subfigure}
    \caption{Examples of the proposed $\epsilon$-insensitive losses defined on $\mathbb R^2$ for different values of $p$.}
    \label{fig:epslosses}
    \end{center}
\end{figure}

Then in Fig.~\ref{fig:epslosses} we highlight the influence of $p$ on the shape of the $\epsilon$-insensitive loss $\ell_\epsilon^p$ defined on $\mathbb R^2$. We set $\epsilon = 1$ and consider values of $p \in \{1.01, 1.5, 2, 3, 5, + \infty \}$. We display $\ell_\epsilon^{1.01}$ in Fig.~\ref{fig:eps1012d}, $\ell_\epsilon^{1.5}$ in Fig.~\ref{fig:eps152d}, $\ell_\epsilon^{2}$ in Fig.~\ref{fig:eps22d}, $\ell_\epsilon^{3}$ in Fig.~\ref{fig:eps32d}, $\ell_\epsilon^{5}$ in Fig.~\ref{fig:eps52d} and $\ell_\epsilon^{\infty}$ in Fig.~\ref{fig:epsinf2d}. 

\begin{figure}
     \begin{center}
     \begin{subfigure}[b]{0.47\textwidth}
        \centering
        \includegraphics[width=\linewidth]{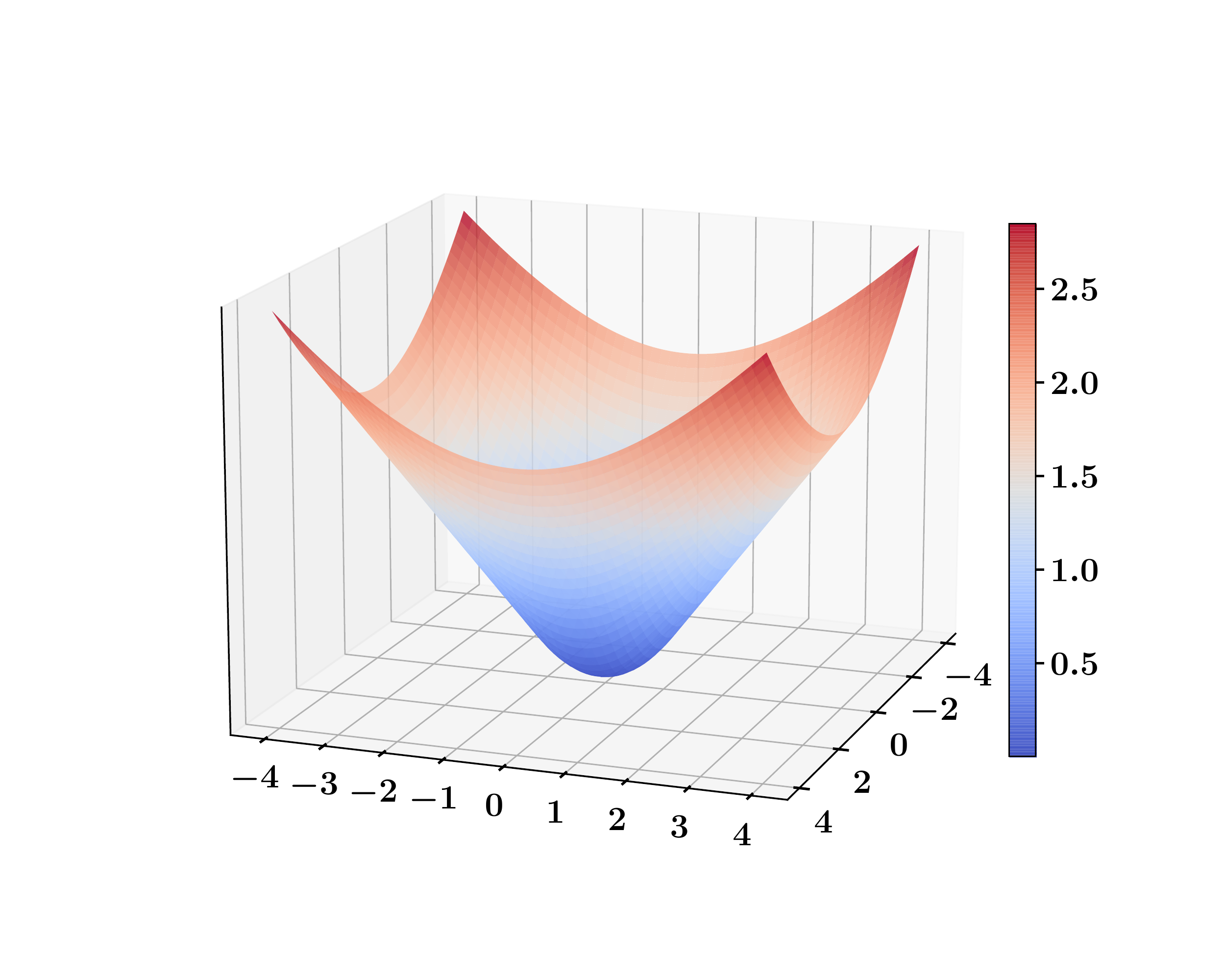}
	    \caption{$H_\kappa^2$ ($\kappa=0.8$)}
	    \label{fig:hub22d}
     \end{subfigure}
     \begin{subfigure}[b]{0.47\textwidth}
        \centering
        \includegraphics[width=\linewidth]{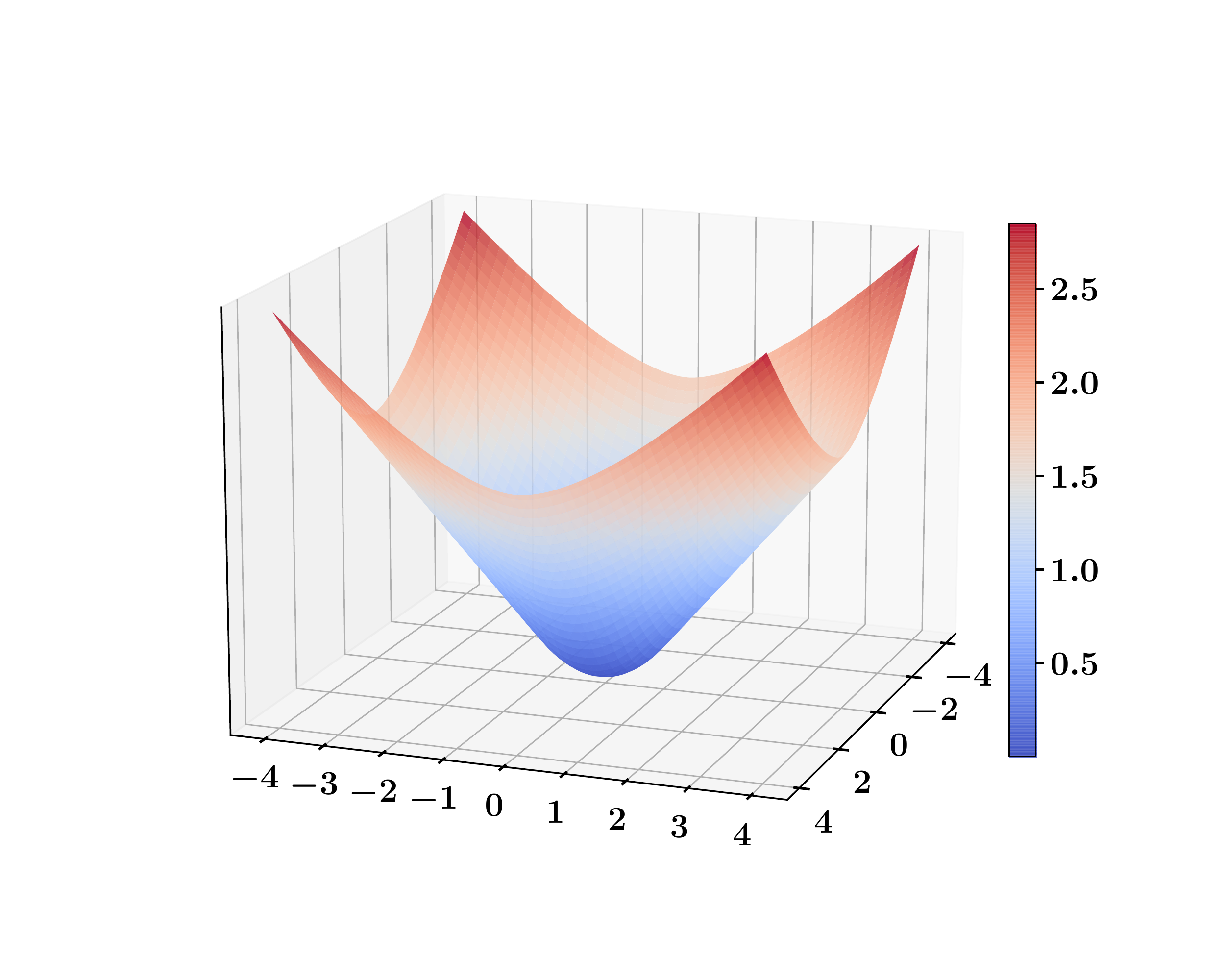}
        \caption{$H_\kappa^{1.5}$ ($\kappa=0.8$)}
        \label{fig:hub152d}
     \end{subfigure}
      \begin{subfigure}[b]{0.47\textwidth}
        \centering
        \includegraphics[width=\linewidth]{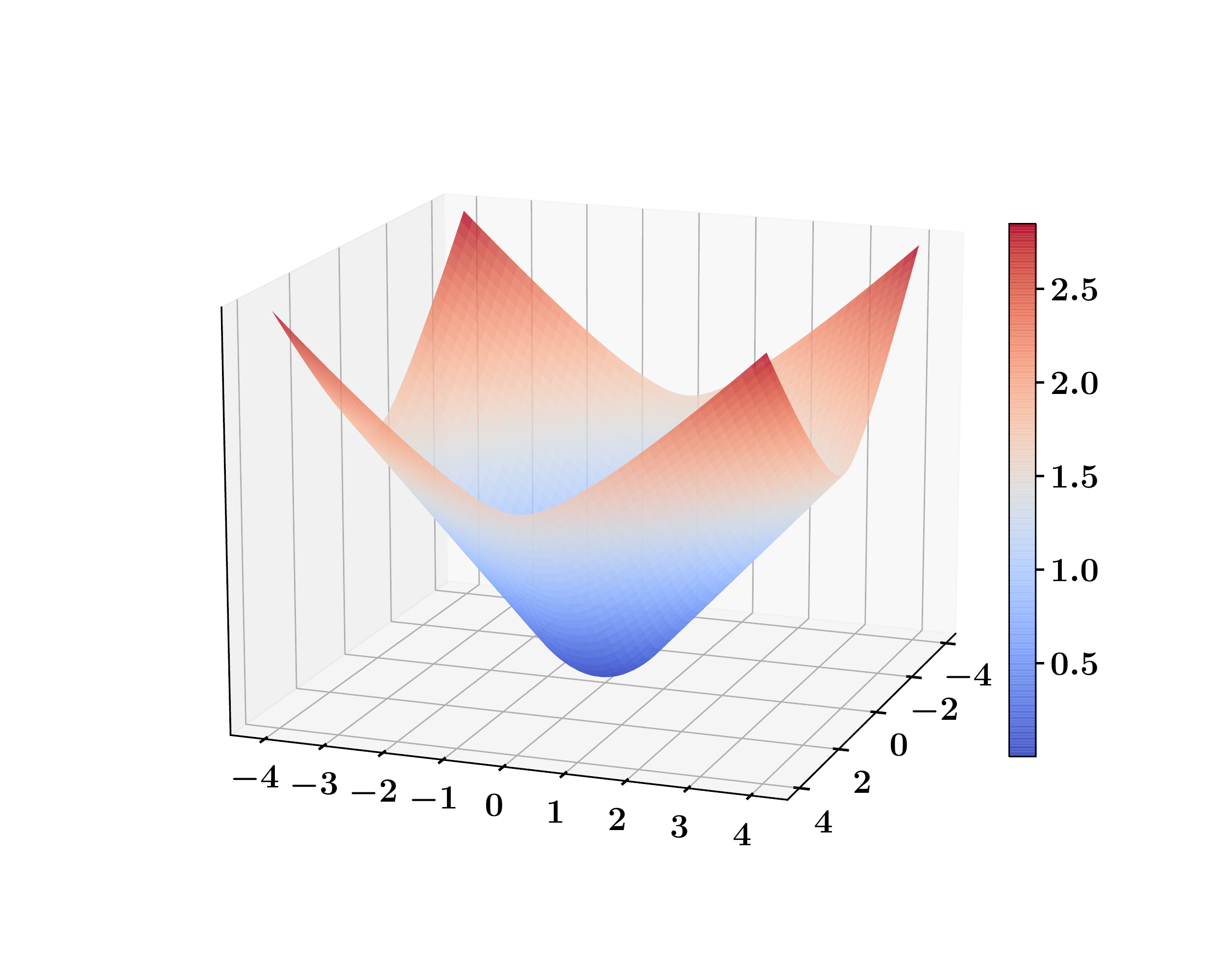}
        \caption{$H_\kappa^{1.25}$ ($\kappa=0.8$)}
        \label{fig:hub1252d}
     \end{subfigure}
     \begin{subfigure}[b]{0.47\textwidth}
        \centering
        \includegraphics[width=\linewidth]{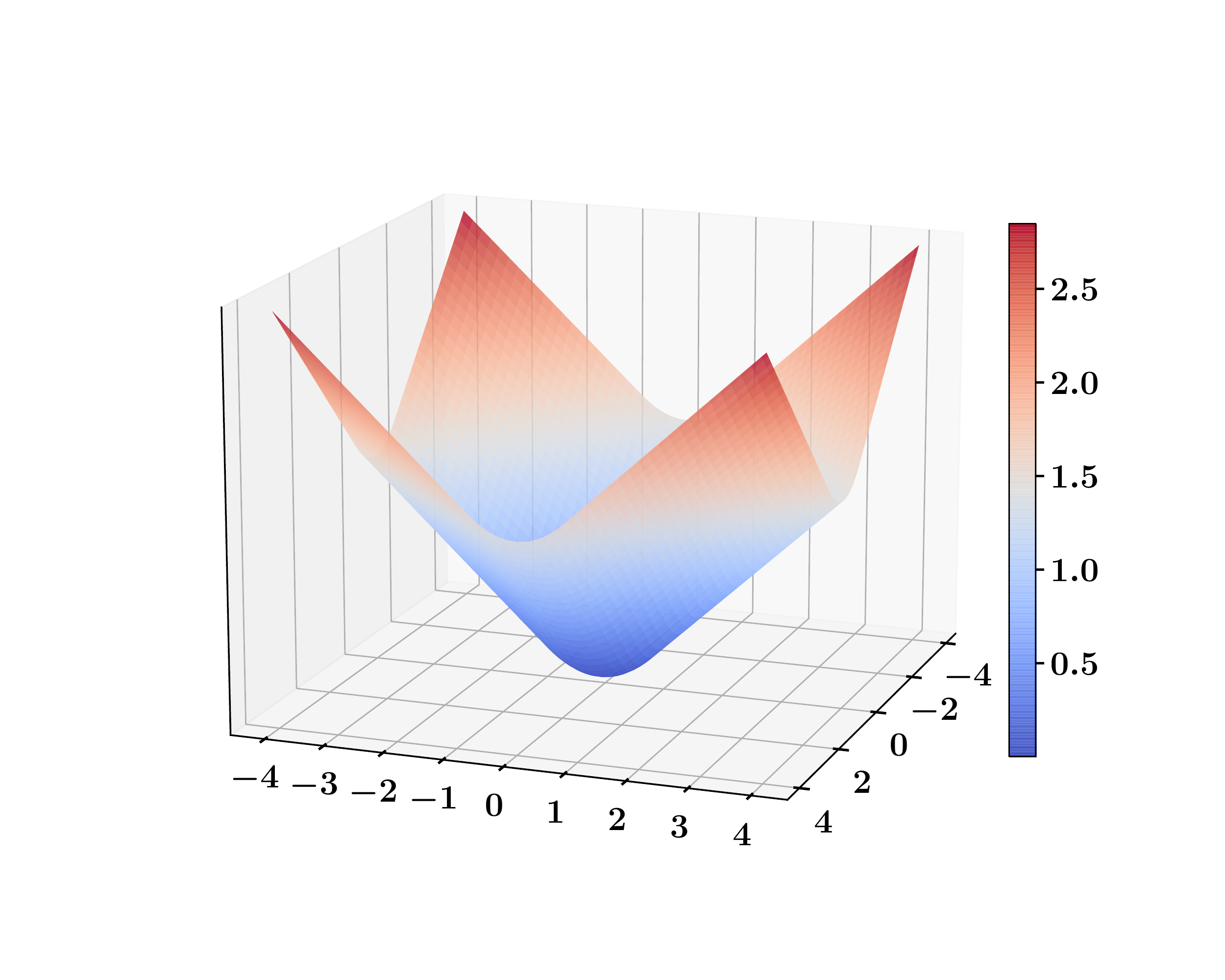}
        \caption{$H_\kappa^1$ ($\kappa=0.8$)}
        \label{fig:hub12d}
     \end{subfigure}
    \caption{Examples of the proposed Huber losses defined on $\mathbb R^2$ for different values of $p$.}
    \label{fig:hublosses}
    \end{center}
\end{figure}

Finally, in Fig.~\ref{fig:hublosses} we underline the influence that the parameter $p$ has on our proposed Huber losses when it is defined on $\mathbb R^2$; we take $\kappa=0.8$ and we display $H_\kappa^{2}$ in Fig.~\ref{fig:hub22d}, $H_\kappa^{1.5}$ in Fig.~\ref{fig:hub152d}, $H_\kappa^{1.25}$ in Fig.~\ref{fig:hub1252d} and $H_\kappa^1$ in Fig.~\ref{fig:hub12d}.

\end{document}